\documentclass[11pt]{article}

\usepackage[final]{acl}

\usepackage{times}
\usepackage{latexsym}
\usepackage{xurl}
\usepackage{hyperref}

\usepackage[T1]{fontenc}

\usepackage[utf8]{inputenc}

\usepackage{microtype}

\usepackage{inconsolata}

\usepackage{graphicx}

\usepackage{hyperref}
\usepackage{booktabs}
\usepackage{rotating}
\usepackage{color}
\usepackage{xcolor}
\usepackage{subcaption}
\usepackage{array}
\usepackage{tcolorbox}
\usepackage{verbatim} 
\usepackage{enumitem}
\usepackage{pifont}
\usepackage{amssymb}
\usepackage{listings}
\usepackage{soul}
\usepackage{float}
\usepackage{amsmath}
\usepackage{multirow}

\usepackage{dsfont}

\definecolor{pastelgreen}{RGB}{102,204,102}
\definecolor{pastelred}{RGB}{255,77,77}
\definecolor{pastelblue}{rgb}{0.68, 0.85, 0.9} 

\newif\ifshownotes
\shownotestrue  

\newcommand{\revision}[1]{\textcolor{black}{#1}}

\DeclareRobustCommand{\DE}[3]{#3}

\title{Evaluating Style-Personalized Text Generation: Challenges and Directions}

\author{Anubhav Jangra$^{1\dagger\ddagger}$, Bahareh Sarrafzadeh$^2$, \textbf{Silviu Cucerzan}$^2$, \\
  \textbf{Adrian de Wynter}$^{2,3*}$, \textbf{Sujay Kumar Jauhar}$^{2\dagger*}$ \\
  $^1$Columbia University, USA \\
  $^2$Microsoft, USA \\
  $^3$The University of York, UK \\
}

\begin{document}
\maketitle
\begingroup
\def\thefootnote{$\dagger$}\footnotetext{ Corresponding authors: \texttt{anubhav@cs.columbia.edu} and \texttt{sjauhar@microsoft.com}.}\def\thefootnote{\arabic{footnote}}
\def\thefootnote{$\ddagger$}\footnotetext{ Work done during an internship at Microsoft.}\def\thefootnote{\arabic{footnote}}
\def\thefootnote{*}\footnotetext{ denotes equal contribution}
\endgroup

\begin{abstract}
With the surge of large language models (LLMs) and their ability to produce customized output, style-personalized text generation--``write like me''--has become a rapidly growing area of interest. 
However, style personalization is highly specific, relative to every user, and depends strongly on the pragmatic context, which makes it uniquely challenging. 
Although prior research has introduced benchmarks and metrics for this area, they tend to be non-standardized and have known limitations (e.g., poor correlation with human subjects). 
LLMs have been found to not capture author-specific style well, it follows that the metrics themselves must be scrutinized carefully. 
In this work we critically examine the effectiveness of the most common metrics used in the field, such as BLEU, embeddings, and LLMs-as-judges. 
We evaluate these metrics using our proposed style discrimination benchmark, which spans eight diverse writing tasks across three evaluation settings: \textit{domain discrimination}, \textit{authorship attribution}, and \textit{LLM-generated personalized vs non-personalized discrimination}. 
We find strong evidence that employing ensembles of diverse evaluation metrics consistently outperforms single-evaluator methods, and conclude by providing guidance on how to reliably assess style-personalized text generation.\footnote{Code and data:  \url{https://github.com/anubhav-jangra/style-personalized-text-evaluation}.}
\end{abstract}

 \section{Introduction}

Large language models have many capabilities that have powered their end-user adoption. 
One core feature is their ability to perform writing assistance in multiple domains, such as journalism \cite{diakopoulos2019automating}; legal services \cite{magesh2024hallucination}; academic writing \cite{khalifa2024using, nguyen2024human}; to name a few.
They have also gained traction in more personal tasks, including drafting emails \cite{li2025emails}, résumés and cover letters \cite{zinjad2024resumeflow}, and social media posts \cite{long2023tweetorial, jain2023ai}. 
What these applications have in common is their need to have some level of style personalization; and, in particular, style-personalized text generation (SPTG; \citealt{mysore2025prototypical}). 
Even though SPTG is neither new nor exclusive to LLMs, it has not been until the latter's advent that this technology has been ubiquitous. 

SPTG evaluation is complex: it is susceptible to user preferences, specificity, and shifts in the pragmatic context. 
Contemporary SPTG is even more challenging; thanks to the capabilities of LLMs, it is now \textit{long-form} and \textit{open-ended}. 
There is also a scarcity of high-quality, high-volume datasets. 
All of this, in turn, makes SPTG notoriously hard to evaluate. 
In response, the research community has pushed to develop new metrics \cite{li2024learning, alhafni2024personalized} and benchmarks \cite{salemi2023lamp, kumar2024longlamp}. 
However, these measurement practices are disjoint and lack standardization. Existing works rely on metrics such as n-gram overlap metrics (e.g., \texttt{BLEU} \cite{papineni2002bleu}), embeddings, and LLMs-as-judges \cite{gu2024survey, li2024llms}, even though they have known limitations (e.g, \citealt{reiter-2018-structured,de-wynter-2025-awes}). 
Since LLMs themselves are known to be ineffective at emulating author-specific content \cite{bhandarkar2024emulating}, it raises the question as to whether the metrics used were truly measuring what was intended; or, better yet, how can the measurement process in SPTG be improved. 
For this, there is a need to evaluate \textbf{to what extent these metrics are effective at measuring contemporary SPTG}. 

In this work, we evaluate the style discrimination capabilities of various metrics alongside two newer paradigms: style-aware embeddings \cite{wegmann2022same, patel2024styledistance} and LLM-as-judge approaches. 
Our focus is on open-ended, long-form SPTG. 
To resemble real-world conditions, we define a low-resource style discrimination setting, where a limited amount of reference style text (fewer than 1,500 words) is available. 
Our analysis spans eight diverse writing tasks and three evaluation settings, including ensembles of these metrics. 
To our knowledge, our work is the first to critically evaluate common and contemporary style-personalized text generation (SPTG) metrics. 
Our main contributions are as follows:
\revision{
\begin{enumerate}[itemsep=0pt, parsep=0pt, topsep=0pt, partopsep=0pt]
    \item We introduce a diverse evaluation benchmark spanning eight diverse writing tasks and three evaluation settings to rigorously assess the style discrimination task. 
    \item We conduct a comprehensive assessment of three major metric paradigms for SPTG: ngram-based metrics, style-embeddings, and LLM-as-judges, highlighting their respective strengths and limitations.
    \item We provide empirical evidence that ensembles of weak evaluation metrics can produce more robust judgments than any individual metric alone.
    \item We analyze the practical challenges of obtaining high-quality human annotations for SPTG evaluation. 
\end{enumerate}
}

\begin{figure*}[h]
	\centering
	\includegraphics[width=\linewidth]{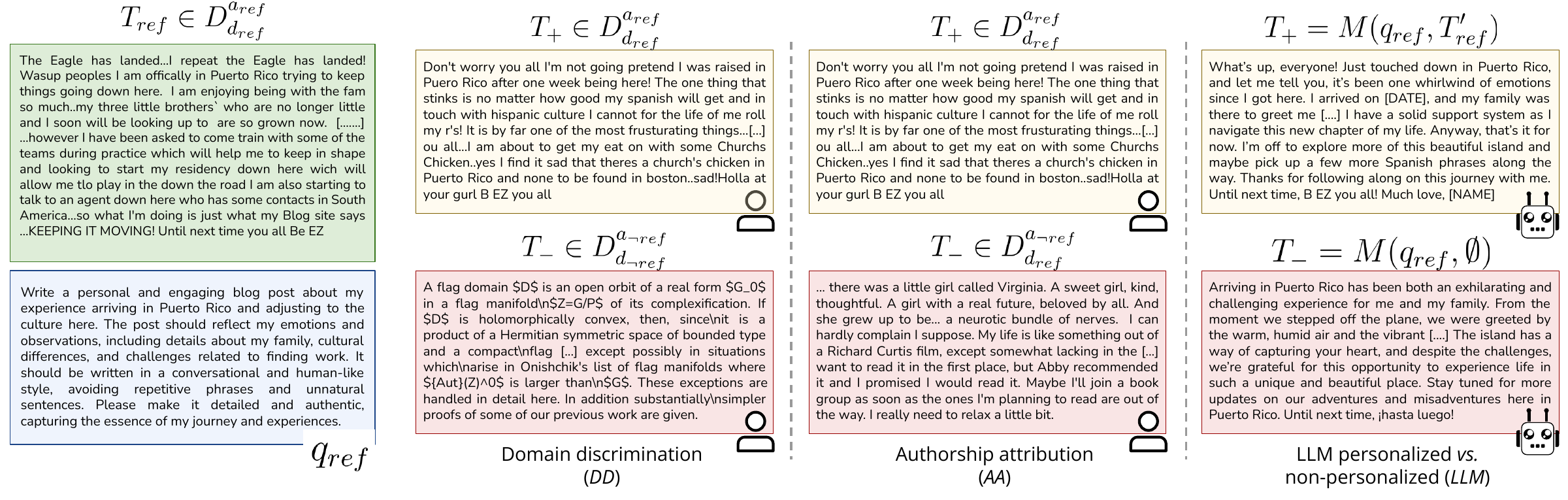}
	\caption{\revision{Illustration of the style discrimination task. Given a reference text  $T_{ref} \in D^{a_{ref}}_{d_{ref}}$ (texts written by author $a_{ref}$ in domain $d_{ref}$), the goal is to identify which candidate text ($T_+$ or $T_-$) is stylistically closer. The three panels show evaluation settings obtained by varying author/domain constraints: domain discrimination (\textit{DD}), authorship attribution (\textit{AA}), and LLM generation, where candidates are produced by a model $M$ with or without style grounding, using a reconstructed user query $q_{ref}$.}}
	\label{fig:intro_fig}
\end{figure*}

\section{Background} \label{sec:related_work}

\subsection{Traditional (non-LLM) measurements}
SPTG builds upon early work in text style transfer. 
For a comprehensive overview of this area, see the surveys by \citet{hu2022text,mukherjee2024survey}; and \citet{liang2024controllable}. 
Text style transfer is usually evaluated in three dimensions: fluency, style accuracy, and content preservation \cite{briakou2021review}. 
There are a variety of metrics to assess style accuracy, ranging from simple n-gram overlap measures (e.g. \texttt{BLEU}; \texttt{ROUGE}, \citealt{lin2004rouge}; \texttt{METEOR}, \citealt{banerjee2005meteor}), to more sophisticated approaches. 
The most common, non-LLM-based metrics are the Wasserstein distance over style-specific lexicons \cite{pele2009fast,mir2019evaluating} and style-specific classifier scores \cite{fu2018style, krishna2020reformulating, jangra2022t}. 
Most of these efforts focused on sentence-level style transfer, which \textbf{does not align with the demands of contemporary SPTG} in long-form, low-resource scenarios \revision{for two main reasons: (a) they typically require larger parallel or pseudo-parallel datasets, which are not available in contemporary SPTG tasks; and (b) their style assessments are designed for short, isolated sentences and therefore cannot capture higher-level stylistic phenomena critical for long-form style discrimination, such as discourse structure, narrative coherence, and co-reference style.}

More generally, it is crucial to note that \textbf{n-gram overlap measures have several shortcomings}, such as not being fully applicable to text generation \cite{liu-etal-2016-evaluate}, nor correlating well with human judgments \cite{gehrmann2023repairing,reiter-2018-structured}. 
\revision{These issues also exist in style transfer, and range from being prone to exploitation--for example, by sampling random sentences from the target distribution to inflate style-accuracy scores \cite{krishna2020reformulating}--to having weak correlations with human judgments \cite{pang2018unsupervised}.}
However, \textbf{ensembling various metrics has been found to improve performance}, at least in machine translation \cite{vstefanik2021regressive}. 
In our work, we integrate this latter finding within our evaluations.

\subsection{LLM-based generation and measurement}

LLMs can perform long-form text generation from minimal grounding data, which makes measurement of LLM-based systems challenging, chiefly because of their sensitivity to the prompt \cite{de-wynter-2025-awes}. 
\revision{\textbf{LLM-based SPTG} is no exception; however, its evaluation is further complicated by the coexistence of \textbf{legacy n-gram metrics and contemporary LLM-as-judge approaches}, which were designed under different assumptions about how generation quality should be measured, especially for long-form text.}
Works relying on n-gram-based methods are ubiquitous. 
For example, previous work applying LLMs to SPTG in prompt rewriting \cite{li2024learning}; steerage through inference \citep{zhang2025personalized}; and retrieval-augmented personalization \citep{mysore2023pearl} were measured with (in order) \texttt{BLEU} and \texttt{ROUGE-1,2,L}; \texttt{ROUGE-1,2} and \texttt{BertScore-F1} \cite{zhang2019bertscore}; and \texttt{ROUGE-L} and \texttt{METEOR}. 
Benchmarks such as LaMP \cite{salemi2024lamp} and LongLaMP \cite{kumar2024longlamp} also use \texttt{ROUGE-1,L}; with \texttt{METEOR} added to the latter. 
Alternate approaches use embedding-based approaches \cite{alhafni2024personalized,horvitz2024tinystyler,shang2019semi}; policy optimization \cite{liu2024authorship}; 
and LLMs-as-judges \cite{ostheimer2023text, lai2023multidimensional, logacheva2022study}. 
Nonetheless, an evaluation by \citet{bhandarkar2024emulating} revealed \textbf{significant limitations on an LLM's ability to emulate author-specific styles, thus suggesting that the measurements above were insufficient}. Their own evaluation metrics were five variants on authorship attribution classifiers in \textit{high-resource} texts (at least 50,000 words per author), \revision{and cannot be adapted to the low-resource SPTG setting in our work, which uses 100-1000 words for style discrimination.}

Extending metrics to long-form SPTG is non-trivial, 
\revision{as several evaluation dimensions commonly used in text style transfer become less informative in the context of modern LLMs 
For instance, fluency is no longer a meaningful discriminative dimension, as contemporary LLM-generated outputs in our setting are generally fluent. Similarly, content preservation has been extensively studied in prior work \cite{lyu2024towards} and is not the primary challenge in contemporary SPTG.}
On the other hand, the difficulty of measuring LLMs, the limitations around known metrics, and the findings on the capabilities of LLMs to emulate style call for a deeper understanding of \textit{what} is measured in SPTG. We take a pragmatic approach to it and focus solely on style accuracy. 

\begin{table*}[]
\caption{Statistics for our extended style personalization evaluation benchmark across the three evaluation settings - \textit{domain discrimination}, \textit{authorship attribution}, and \textit{LLM-generated personalized vs non-personalized}, with corresponding columns describing the average number of words in the $T_{ref}$, $T_+$, and $T_-$ respectively.} \label{tab:our_data}
\centering
\resizebox{0.8\textwidth}{!}{%
\begin{tabular}{lcccc}
\toprule
\textbf{Dataset Name} & \textbf{\# Authors} & \textbf{\textit{Domain Discrimination}} & \textbf{\textit{Authorship Attribution}} & \textbf{\textit{LLM Generated}} \\
\midrule
\textbf{\textit{Amazon}}  & 100 & 157/267/388 & 161/267/263  & 267/272/300   \\
\textbf{\textit{ArXiV}}  & 100 & 144/224/447 & 144/224/229  & 224/406/505   \\
\textbf{\textit{Blogs}} & 95 & 173/329/373 & 164/329/349 & 329/369/396  \\
\textbf{\textit{Enron}}  & 100 & 123/328/520 & 119/328/334  & 328/193/219  \\
\textbf{\textit{Lyrics}}  & 95 & 210/278/495 & 211/278/288  & 278/308/315   \\
\textbf{\textit{Reddit}} & 56  & 87/38/410 & 93/38/38   & 38/222/261    \\
\textbf{\textit{Reuters}} & 49  & 363/506/461 & 366/506/507 & 506/416/456   \\
\textbf{\textit{Short Stories}} & 41  & 903/1289/267 & 911/1289/1364 & 1289/543/515 \\
\bottomrule
\end{tabular}%
}
\end{table*}

\section{Problem Statement}

For an author $a$ who wrote texts in the domain $d$, let $D^a_d$ denote the set of said texts. 
Then, given a reference text $T_{ref} \in D^{a_{ref}}_{d_{ref}}$, and two candidate texts $T_+ \in D^{a_1}_{d_1}$ and $T_- \in D^{a_2}_{d_2}$, the \textit{style discrimination task} aims to identify which candidate is stylistically closer to the reference. 
The choice of $a_1, d_1$ and $a_2, d_2$ alters the evaluation setting (e.g., authorship attribution will consider $d_1 = d_2$ but $a_1 \neq a_2$), and 
are defined in Section~\ref{sec:exp_setting}.
Assuming an underlying style distribution $S_x$ for each text $T_x$, such that $S_{ref} \approx S_+$ and $S_{ref} \not\approx S_-$, the goal is to learn a discriminator $f : (T_{ref}, T_+, T_-) \mapsto \{+, -\}$ that predicts whether $T_+$ or $T_-$ is more similar in style to $T_{ref}$. 
This framing is thus a binary classification problem \revision{to enable robust style measurement}.
Rather than requiring annotators or metrics to assign an arbitrary degree of stylistic similarity, we use comparative judgments, which have been shown to yield more reliable judgments than absolute ratings \cite{goffin2011all}.
Hence, we reduce this problem to determining if the metrics \textit{can} measure style accuracy, rather than to what extent. 
\section{Experimental Settings} \label{sec:exp_setting}

\subsection{Datasets} \label{subsec:dataset}

\paragraph{Source dataset selection.} 
We selected eight publicly available NLP datasets encompassing diverse writing tasks and domains. 
They were chosen based on the following criteria: (i) they must include multiple writing instances produced by the same author, (ii) they must represent realistic and well-defined writing tasks, and (iii) they must be openly accessible and licensed for research use. 
The datasets span diverse writing tasks, namely, creative writing (\textit{short stories}, \citealt{james2019stories}; \textit{lyrics}, \citealt{edenbd_150k_lyrics_2021}) formal writing (\textit{Reuters news}, \citealt{reuters-21578_text_categorization_collection_137}; \textit{Enron emails}, \citealt{klimt2004introducing}; \textit{ArXiv scientific abstracts}, \citealt{arxiv_org_submitters_2024}), and informal writing (\textit{blogs}, \citealt{schler2006effects}; \textit{Amazon food reviews}, \citealt{mcauley2013amateurs}; \textit{Reddit microblogs}, \citealt{patel2022low}).


\paragraph{Domain-level down-sampling.} We randomly sampled instances from each dataset based on the number of tokens written by each author. We used the  \texttt{tiktoken} library\footnote{\url{https://github.com/openai/tiktoken}} to determine the number of tokens in each article. For each author $a$ in domain $d$, we build $\mathcal{D}^a_d$ by randomly sampling the articles written by the author. 
We set an upper limit of 500 tokens \revision{for the style reference text ($T_{ref}$)}, and randomly selected target articles with 200 to 700 tokens. 
\textit{Reddit} and \textit{short stories} domains had different token lengths due to their non-standard text lengths. 
\textit{Reddit} domain instances were too short, and hence we removed the lower limit for the target article selection. 
The instances in the \textit{short stories} domain were too long, and we increased the upper limit of grounding articles to 1500 tokens and target articles to 2000 tokens. 
After this sampling step, we obtained 50 instances for \textit{Reddit}, \textit{Reuters}, and \textit{short stories} domains, and 100 instances for the remaining domains. 
We removed a few examples from the dataset where the personalized text generation model did not yield an appropriate response\footnote{\revision{We removed a few examples (specifically, 5 from \textit{blogs}, 5 from \textit{lyrics}, 1 from \textit{Reuters} and 1 from \textit{short stories}) because the generation model used to generate $T_+$ or $T_-$ for the LLM evaluation setting did not generate responses due to its guardrailed output for plausible copyrighted input.}}, leading to a final dataset comprised of 636 instances. 
Table \ref{tab:our_data} summarizes the overall statistics of the evaluation benchmark.

\subsection{Evaluation settings} \label{subsec:eval_setting}
To assess style discrimination under different conditions \revision{(see Figure~\ref{fig:intro_fig} for example)}, we define three evaluation settings based on the source of the candidate set $\{T_+, T_-\}$. For a given reference text $T_{ref} \in D^{a_{ref}}_{d_{ref}}$, the candidates $T_+$ and $T_-$ are randomly sampled as follows- 

\paragraph{Domain discrimination (\textit{DD}):}\footnote{Domain discrimination is primarily a sanity check to contextualize results in the other settings. While it does not capture the nuances of real-world style evaluation, success on this condition is necessary for a metric to meaningfully evaluate stylistic differences.} sample $T_+ \sim D^{a_{ref}}_{d_{ref}}$ with $T_+ \neq T_{ref}$ and $T_{-} \sim D^{a_.}_{d_{\neg ref}}$. 

\paragraph{Authorship attribution (\textit{AA}):} sample $T_+ \sim D^{a_{ref}}_{d_{ref}}$ with $T_+ \neq T_{ref}$ and $T_{-} \sim D^{a_{\neg ref}}_{d_{ref}}$.

\paragraph{LLM personalized versus non-personalized (\textit{LLM}):} suppose a generative model maps an author's queries (prompts) and sample texts to output texts, $M \colon \mathbf{Q}^{a}_{d} \times D_{d'}^{a'} \rightarrow D_{d}^{a}$. 
Then a query reconstruction function $f_{query} : D_{d}^a\rightarrow \mathbf{Q}^{a}_d$ is an approximate inverse function $f_{query} \approx  M^{-1}$ that takes in outputs and returns queries. $f_{query}$ was implemented using an LLM as a meta-prompting approach \cite{dewynter2023metaprompting}. To resemble human-written queries, we used queries from \texttt{WildChat} \cite{zhao2024wildchat}. 
For that, we randomly selected 50,000 English-language user interactions and performed two consecutive filtering steps using text classification: \revision{i) discarding all the user prompts that do not correspond to writing requests, and ii) discarding the writing prompts of irrelevant writing tasks for our dataset domains.}
The filtered user queries are used to generate the reconstructed user query $q_{ref}$. 
Using a generation model $M$, we then generate $T_+ = M(q_{ref}, T'_{ref})$ and $T_- = M(q_{ref}, \emptyset)$, where $T'_{ref} \sim D^{a_{ref}}_{d_{ref}}$ such that $T'_{ref} \neq T_{ref}$\footnote{Since $T_+$ is generated by an LLM, it may not fully capture the target author's style, introducing additional variability into the evaluation. Nevertheless, this setting is integral, as it closely reflects the practical use of personalized generation systems, where stylistic fidelity must be evaluated on model-generated rather than human-authored outputs.}. 
See Appendices~\ref{app:callparameters} for detailed call parameters and~\ref{app:prompts} for the prompts used.

\subsection{Evaluation metrics} \label{subsec:baselines}



\paragraph{Random 100.} The average accuracy of 100 random binary classifier iterations (expected baseline: 50.0\%).

\paragraph{N-gram overlap-based evaluation metrics.} We explore three widely adopted n-gram evaluation metrics: \texttt{BLEU}, \texttt{ROUGE}, and \texttt{METEOR}. For a n-gram similarity measure $Sim \colon D_{d}^a \times D_{d'}^{a'} \rightarrow [0, 1]$, we obtain the binary label as $\mathds{1}\left[Sim(T_{ref}, T_+) > Sim(T_{ref}, T_-)\right]$.

\paragraph{Style Embedding-based evaluation metrics.} We evaluate \texttt{Wegmann}\footnote{For simplicity, we refer to this baseline as ``Wegmann'' after the first author in \citet{wegmann2022same}.} \cite{wegmann2022same}, and \texttt{StyleDistance} \cite{patel2024styledistance} style embeddings. Using cosine similarity between the embeddings, we assign the binary label as $\mathds{1}[Sim_{cos}(\vec{T_{ref}}, \vec{T_+)} > Sim_{cos}(\vec{T_{ref}}, \vec{T_-})]$.


\paragraph{LLM-as-judge evaluation metrics.} 
We evaluate a range of open-source models, including \texttt{Ministral-3B}, \texttt{Llama-3.1-8B} \cite{grattafiori2024llama}, \texttt{Mistral-24B}, \texttt{Qwen3-32B} \cite{yang2025qwen3}, \texttt{DeepSeek-V3}\footnote{We use \texttt{Mistral-24B} and \texttt{DeepSeek-V3} to refer to \texttt{Mistral-Small-24B-Instruct-2501} and \texttt{DeepSeek-V3-0324} models respectively.} \cite{liu2024deepseek}. We also evaluate OpenAI's closed-source models \texttt{o4-mini} and \texttt{gpt-4.1} \cite{hurst2024gpt}, covering multiple model families and parameter sizes. 
See Appendices~\ref{app:prompts} and~\ref{app:callparameters} for the evaluation prompt and call parameters, respectively. 


\subsection{Ensemble of evaluation metrics}

We evaluate several ensembles of evaluation metrics as follows:
\begin{itemize}
    \itemsep-0.5em
    \item $\rho_{ngram}$: ensembling over n-gram evaluation metrics.
    \item $\rho_{\neg LLM}$: ensembling over all non-LLM evaluation metrics.
    \item $\rho_{LLM}$: ensembling over LLM-as-judge evaluation metrics.
    \item $\rho_{OS}$: ensembling over all open-source evaluation metrics.
    \item $\rho_{all}$: ensembling over all evaluation metrics detailed in Section \ref{subsec:baselines}.
\end{itemize}
\noindent 
We explore two ensembling strategies: majority voting ($MV$) and performance-weighted voting ($PWV$). \revision{$PWV$ ensemble strategy was implemented by using the normalized aggregate of individual metric scores with a threshold of 0.5 to obtain the binary decision\footnote{$PWV$ uses weights proportional to each metric's standalone performance. We use this as an oracle-style upper bound to estimate the potential benefit of weighting metrics by their reliability, rather than as a deployable ensemble whose weights are tuned on target data.}.}

\paragraph{Statistical Significance}
All accuracies are reported with 95\% confidence intervals from paired bootstrap resampling of the evaluation set (10,000 resamples). Differences between metrics are tested with McNemar's test \cite{mcnemar1947note} on their paired binary decisions, Holm–Bonferroni corrected for the number of comparisons within each column.

\section{Results} \label{sec:results}

\begin{table*}[t]
\caption{Accuracy of evaluation metrics for \textit{DD}, \textit{AA}, and \textit{LLM}. Scores in \textbf{bold} and \underline{underline} indicate the best and the second best metrics for each evaluation paradigm respectively. Subscripts are 95\% confidence intervals from paired bootstrap resampling (10,000 resamples). $\dagger$ marks results that are not significantly different from the best result in that column.} \label{tab:base_results}
\centering
\small
\begin{tabular}{lcccc}
\toprule
\textbf{Evaluation Metric} & \textit{\textbf{DD}} & \textit{\textbf{AA}} & \textit{\textbf{LLM}} & Mean \\
\midrule
\texttt{Random} 100 & 0.500$_{\pm 0.048}$ & 0.500$_{\pm 0.048}$ & 0.500$_{\pm 0.048}$ & 0.500$_{\pm 0.048}$ \\
\hline
\texttt{BLEU} & \textbf{0.869$_{\pm 0.027}$} & \textbf{0.700$_{\pm 0.036}$} & \underline{0.631$_{\pm 0.038}$$^{\dagger}$} & \textbf{0.733$_{\pm 0.022}$} \\
\texttt{ROUGE-1} & \underline{0.858$_{\pm 0.027}$} & \underline{0.695$_{\pm 0.035}$} & 0.613$_{\pm 0.038}$ & \underline{0.722$_{\pm 0.022}$} \\
\texttt{ROUGE-2} & 0.803$_{\pm 0.031}$ & 0.643$_{\pm 0.038}$ & \textbf{0.634$_{\pm 0.038}$$^{\dagger}$} & 0.693$_{\pm 0.024}$ \\
\texttt{ROUGE-L} & 0.854$_{\pm 0.028}$ & 0.673$_{\pm 0.036}$ & \underline{0.631$_{\pm 0.038}$$^{\dagger}$} & 0.719$_{\pm 0.022}$ \\
\texttt{METEOR} & 0.783$_{\pm 0.033}$ & 0.654$_{\pm 0.038}$ & 0.547$_{\pm 0.039}$ & 0.661$_{\pm 0.023}$ \\
\hline
\texttt{Wegmann} & \underline{0.910$_{\pm 0.024}$} & \textbf{0.670$_{\pm 0.038}$} & \underline{0.558$_{\pm 0.039}$} & \underline{0.713$_{\pm 0.020}$} \\
\texttt{StyleDistance} & \textbf{0.926$_{\pm 0.020}$} & \underline{0.665$_{\pm 0.036}$} & \textbf{0.575$_{\pm 0.039}$} & \textbf{0.722$_{\pm 0.021}$} \\
\hline
\texttt{Ministral-3B} & 0.285$_{\pm 0.036}$ & 0.272$_{\pm 0.035}$ & 0.417$_{\pm 0.038}$ & 0.324$_{\pm 0.021}$ \\
\texttt{Llama-3.1-8B} & 0.157$_{\pm 0.028}$ & 0.263$_{\pm 0.035}$ & 0.418$_{\pm 0.039}$ & 0.279$_{\pm 0.019}$ \\
\texttt{Mistral-24B} & 0.478$_{\pm 0.039}$ & 0.453$_{\pm 0.039}$ & 0.470$_{\pm 0.039}$ & 0.467$_{\pm 0.026}$ \\
\texttt{Qwen3-32B} & 0.689$_{\pm 0.036}$ & 0.608$_{\pm 0.038}$ & 0.525$_{\pm 0.039}$ & 0.607$_{\pm 0.024}$ \\
\texttt{DeepSeek-V3} & 0.693$_{\pm 0.036}$ & 0.679$_{\pm 0.036}$ & \underline{0.601$_{\pm 0.038}$} & 0.658$_{\pm 0.024}$ \\
\texttt{o4-mini} & \underline{0.917$_{\pm 0.022}$} & \underline{0.750$_{\pm 0.035}$} & 0.597$_{\pm 0.038}$ & \underline{0.755$_{\pm 0.021}$} \\
\texttt{gpt-4.1} & \textbf{0.961$_{\pm 0.016}$} & \textbf{0.807$_{\pm 0.031}$} & \textbf{0.678$_{\pm 0.036}$} & \textbf{0.815$_{\pm 0.018}$} \\
\bottomrule
\end{tabular}
\end{table*}


\begin{figure*}[t]
	\centering
	\includegraphics[width=0.75\linewidth]{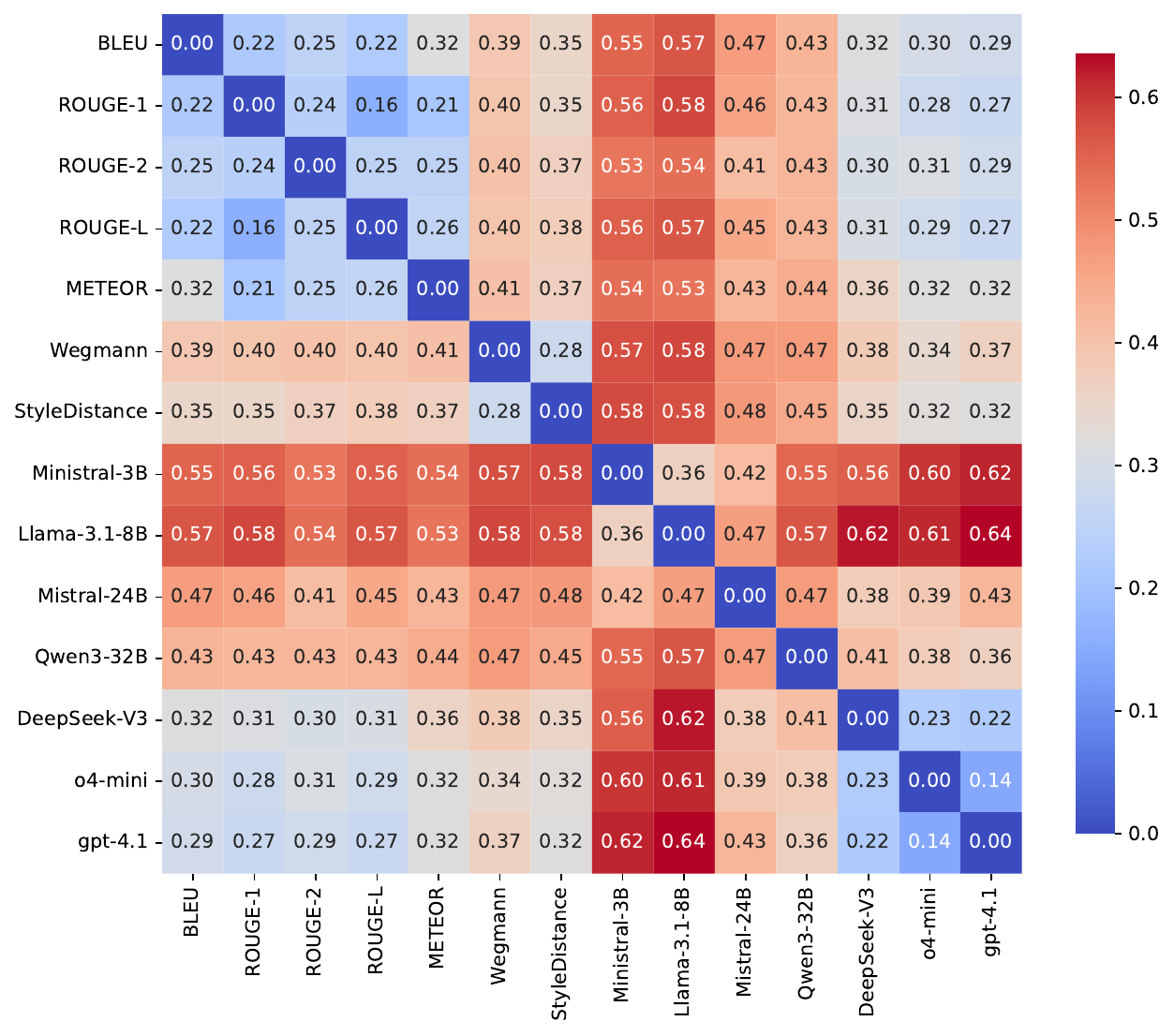}
	\caption{Pairwise disagreement of evaluation metrics for the \textit{authorship attribution} evaluation setting. It can be seen from the figure that evaluation metrics across different evaluation paradigms have higher disagreement compared to metrics within the same evaluation paradigms. For example, even though \texttt{ROUGE-1} and \texttt{StyleDistance} achieve the same overall performance score of 0.722 (see Table \ref{tab:base_results}); their disagreement score is 0.35, compared to the lower disagreement scores of \texttt{ROUGE-1} and \texttt{BLEU} (0.22) and \texttt{StyleDistance} and \texttt{Wegmann} (0.28).}
	\label{fig:disagreement}
\end{figure*}

\subsection{Performance of individual evaluation metrics}

Table \ref{tab:base_results} summarizes the performance of individual evaluation metrics on our proposed evaluation benchmark. We observe notable trends in metric performance across different evaluation strategies and paradigms, which we discuss next.

\paragraph{Performance across evaluation settings} Across nearly all evaluation metrics, performance was highest in the \textit{DD} setting, followed by \textit{AA}, and lowest in \textit{LLM}. 
Over four metrics achieved over 0.9 accuracy in \textit{DD}, with the highest performance of 0.961 accuracy by \texttt{gpt-4.1}. 
Comparatively, \texttt{gpt-4.1} achieved 0.807 accuracy for \textit{AA} and 0.678 accuracy in \textit{LLM}. On average, accuracy dropped by -16.2\% from \textit{DD} to \textit{AA}, and -7.5\% from \textit{AA} to \textit{LLM}.

\paragraph{Performance across evaluation metrics.} 
Close-sourced LLM-as-judge metrics achieved the highest accuracy scores, with \texttt{gpt-4.1} attaining the highest overall accuracy (0.815), followed by \texttt{o4-mini} (0.755). Comparatively, the best performing n-gram metric, \texttt{BLEU}, achieved 0.733 accuracy, and the best performing style embedding metric, \texttt{StyleDistance}, achieved 0.722 accuracy. This ordering, however, is statistically reliable only in the two easier settings, as \texttt{gpt-4.1} is significantly the strongest judge under \textit{DD} and \textit{AA} (all pairwise McNemar $p<0.01$), whereas under \textit{LLM} its lead over \texttt{BLEU}, \texttt{ROUGE-2}, and  \texttt{ROUGE-L} falls within noise ($\Delta \leq 0.047$, $p=0.18$). Across the three evaluation settings, the performance decline from \textit{DD} to \textit{LLM} was the highest for style embedding metrics (-38.3\%), moderate for n-gram-based metrics (-26.7\%), and least for  LLM-as-judge metrics (-11.3\%); even so, the hardest setting is the one in which the cheapest metrics are already competitive with the most expensive.



Model size and access type influenced performance: larger and closed-source language models consistently outperformed smaller models. The three smallest models evaluated had random or near-random performance. See Appendix~\ref{app:llmsresponses} for a distribution of the LLM-as-a-judge responses.




\subsection{Performance of ensemble of metrics}

Table \ref{tab:ensemble_results_main} presents the results of the best-performing ensemble combination across different metrics. Ensembles generally outperformed their individual constituents; for example, the $\rho_{\neg LLM}^{PWV}$ ensemble achieves +6.4\% higher accuracy than its best-performing constituent, \texttt{BLEU}. This benefit, however, is confined to ensembles of comparably weak constituents: $\rho_{all}^{PWV}$ attains the highest mean accuracy overall, yet is not significantly better than \texttt{gpt-4.1} alone in any evaluation setting ($\Delta \leq 0.019$, $p \geq 0.81$), so its gains over its strongest constituent fall within noise. Furthermore, \textit{PWV} ensembles consistently yielded marginal improvements to their \textit{MV} counterparts. The most effective ensembles exhibited a uniform distribution over different evaluation paradigms. See Figure~\ref{fig:disagreement} for a depiction of the pairwise disagreement on evaluation metrics in \textit{AA}. 



\begin{table*}[h]
\caption{Accuracy of best performing ensemble across the three evaluation settings. Scores in \textbf{bold} and \underline{underline} indicate the best and the second best metrics for each evaluation paradigm respectively. Subscripts are 95\% confidence intervals from paired bootstrap resampling (10,000 resamples). $\dagger$ marks results that are not significantly different from the best result in that column.} \label{tab:ensemble_results_main}
\centering
\small
\resizebox{\textwidth}{!}{%
\begin{tabular}{llp{7cm}cccc}
\toprule
 & \textbf{Metric} & \textbf{Ensemble Composition} & \textit{\textbf{DD}} & \textit{\textbf{AA}} & \textit{\textbf{LLM}} & Mean \\
\midrule
\multirow{4}{*}{\rotatebox[origin=c]{90}{Individual}}  & \texttt{Random} 100 & - & 0.500$_{\pm 0.048}$ & 0.500$_{\pm 0.048}$ & 0.500$_{\pm 0.048}$ & 0.500$_{\pm 0.048}$ \\
 & \texttt{BLEU} & - & 0.869$_{\pm 0.027}$ & \underline{0.700$_{\pm 0.036}$} & \underline{0.631$_{\pm 0.038}$} & \underline{0.733$_{\pm 0.022}$} \\
 & \texttt{StyleDistance} & - & \underline{0.926$_{\pm 0.020}$} & 0.665$_{\pm 0.036}$ & 0.575$_{\pm 0.039}$ & 0.722$_{\pm 0.021}$ \\
 & \texttt{gpt-4.1} & - & \textbf{0.961$_{\pm 0.016}$$^{\dagger}$} & \textbf{0.807$_{\pm 0.031}$} & \textbf{0.678$_{\pm 0.036}$$^{\dagger}$} & \textbf{0.815$_{\pm 0.018}$$^{\dagger}$} \\
\midrule
\multirow{5}{*}{\rotatebox[origin=c]{90}{MV}}  & $\rho_{ngram}$ & \texttt{BLEU} \texttt{ROUGE2} \texttt{ROUGEL} & 0.866$_{\pm 0.027}$ & 0.704$_{\pm 0.036}$ & 0.654$_{\pm 0.038}$$^{\dagger}$ & 0.742$_{\pm 0.022}$ \\
 & $\rho_{\neg LLM}$ & \texttt{BLEU} \texttt{ROUGE2} \texttt{ROUGEL} \texttt{Wegmann} \texttt{StyleDistance} & 0.914$_{\pm 0.022}$ & 0.725$_{\pm 0.035}$ & 0.682$_{\pm 0.036}$$^{\dagger}$ & 0.774$_{\pm 0.020}$ \\
 & $\rho_{LLM}$ & \texttt{Qwen3-32B} \texttt{o4-mini} \texttt{gpt-4.1} & \underline{0.937$_{\pm 0.019}$} & \underline{0.781$_{\pm 0.031}$$^{\dagger}$} & 0.640$_{\pm 0.038}$$^{\dagger}$ & \underline{0.786$_{\pm 0.020}$} \\
 & $\rho_{OS}$ & \texttt{BLEU} \texttt{ROUGEL} \texttt{Wegmann} \texttt{StyleDistance} \texttt{DeepSeek-V3} & 0.925$_{\pm 0.020}$ & 0.750$_{\pm 0.033}$ & \underline{0.684$_{\pm 0.036}$$^{\dagger}$} & \underline{0.786$_{\pm 0.019}$} \\
 & $\rho_{all}$ & \texttt{BLEU} \texttt{ROUGEL} \texttt{Wegmann} \texttt{StyleDistance} \texttt{gpt-4.1} & \textbf{0.962$_{\pm 0.016}$$^{\dagger}$} & \textbf{0.786$_{\pm 0.033}$$^{\dagger}$} & \textbf{0.697$_{\pm 0.036}$} & \textbf{0.815$_{\pm 0.018}$$^{\dagger}$} \\
\midrule
\multirow{5}{*}{\rotatebox[origin=c]{90}{PWV}}  & $\rho_{ngram}$ & \texttt{BLEU} \texttt{ROUGE1} \texttt{ROUGE2} \texttt{ROUGEL} & 0.862$_{\pm 0.027}$ & 0.714$_{\pm 0.035}$ & 0.654$_{\pm 0.038}$$^{\dagger}$ & 0.743$_{\pm 0.021}$ \\
 & $\rho_{\neg LLM}$ & \texttt{BLEU} \texttt{ROUGE2} \texttt{Wegmann} \texttt{StyleDistance} & \underline{0.948$_{\pm 0.017}$$^{\dagger}$} & 0.722$_{\pm 0.035}$ & 0.670$_{\pm 0.036}$$^{\dagger}$ & 0.780$_{\pm 0.019}$ \\
 & $\rho_{LLM}$ & \texttt{Qwen3-32B} \texttt{DeepSeek-V3} \texttt{o4-mini} \texttt{gpt-4.1} & 0.945$_{\pm 0.017}$ & \underline{0.772$_{\pm 0.033}$} & 0.662$_{\pm 0.038}$$^{\dagger}$ & \underline{0.793$_{\pm 0.019}$} \\
 & $\rho_{OS}$ & \texttt{BLEU} \texttt{ROUGEL} \texttt{Wegmann} \texttt{StyleDistance} \texttt{Qwen3-32B} \texttt{DeepSeek-V3} & 0.937$_{\pm 0.019}$ & 0.750$_{\pm 0.033}$ & \underline{0.682$_{\pm 0.036}$$^{\dagger}$} & 0.790$_{\pm 0.019}$ \\
 & $\rho_{all}$ & \texttt{BLEU} \texttt{ROUGE1} \texttt{StyleDistance} \texttt{gpt-4.1} & \textbf{0.967$_{\pm 0.014}$} & \textbf{0.802$_{\pm 0.031}$$^{\dagger}$} & \textbf{0.693$_{\pm 0.036}$$^{\dagger}$} & \textbf{0.821$_{\pm 0.018}$} \\
\bottomrule
\end{tabular}
}
\end{table*}

\subsection{Ablation: LLM-as-judges prompting} \label{subsec:apo}

\revision{To examine the robustness of LLM-as-judge as an evaluation metric, we conducted a prompt sensitivity analysis over several prompt formulations. Since LLM-based evaluation can be sensitive to prompt wording, relying on a single prompt risks conflating metric performance with prompt-specific artifacts. Specifically, we further explored three prompt variants: (1) a parseable structured output format ($P_{struct}$), (2) a binary judgment format ($P_{binary}$), and (3) automatically optimized prompts obtained via Automatic Prompt Optimization (APO).}
For APO, we used the strategy proposed by \citet{pryzant2023automatic} to optimize the prompts using natural language gradients. For this we separated out a development set of 64 instances (8 instances per domain), and report the evaluation scores on the remaining 572 instances. We limited this exploration to the \texttt{Llama-3.1-8b} model for all evaluation settings, and leave the exploration over other models to future work. 
The results are in Table~\ref{tab:apo_results}. 
$P_{struct}$ consistently yielded worse performance than the original prompting strategy ($P_{orig}$), with a -24\% decrease in overall performance. 
Comparatively, $P_{binary}$ achieved +34.5\% higher overall accuracy than $P_{orig}$. 
Using APO helped improve the performance of $P_{orig}$ and $P_{struct}$ by +24.5\% and +63.5\%, respectively, while yielding similar performance to $P_{binary}$. See Appendix~\ref{app:prompts} for the prompts used.

\begin{table}[]
\caption{Accuracy of various LLM-as-judge prompting strategies. 
$P_{orig}$, $P_{struct}$, and $P_{binary}$ are the original, structured-parseable, and binary prompts, respectively. $P_X+APO$ is the prompt after APO on $P_X$. 
} \label{tab:apo_results}
\centering
\small
\begin{tabular}{lcccc}
\toprule
\textbf{Prompt Strategy} & \textbf{\textit{DD}}    & \textbf{\textit{AA}}    & \textbf{\textit{LLM}}   & Mean \\\midrule
$P_{orig}$              & 0.279 & 0.383 & 0.476 &    0.379  \\
$P_{orig}+APO$        & \underline{0.504} & 0.463 & 0.448 &    0.472  \\
$P_{struct}$            & 0.160 & 0.238 & 0.467 &    0.288  \\
$P_{struct}+APO$      & 0.448 & 0.477 & 0.488 &    0.471  \\
$P_{binary}$            & \textbf{0.526} & \underline{0.491} & \underline{0.514} &    \textbf{0.510}  \\
$P_{binary}+APO$      & 0.453 & \textbf{0.525} & \textbf{0.539} &    \underline{0.506}  \\
\bottomrule
\end{tabular}
\end{table}

\subsection{Ablation: human agreement}\label{subsec:humanagreement}

Although the focus of our work is automated metrics, we conducted a brief study to evaluate inter-annotator agreement (IAA) in human annotators (Appendix~\ref{app:humanannotation}). 
\revision{The goal was to better understand the relationship of SPTG characteristics across human judgments.}
We found that humans had high IAA as measured by the agreement ratio on content (0.779), moderate-to-high on style (0.641), and low on copy-editing (usefulness of the text; 0.400) preferences, respectively. 

\section{Discussion} \label{sec:discussion}

\begin{figure}[t]
	\centering
	\includegraphics[width=0.85\linewidth]{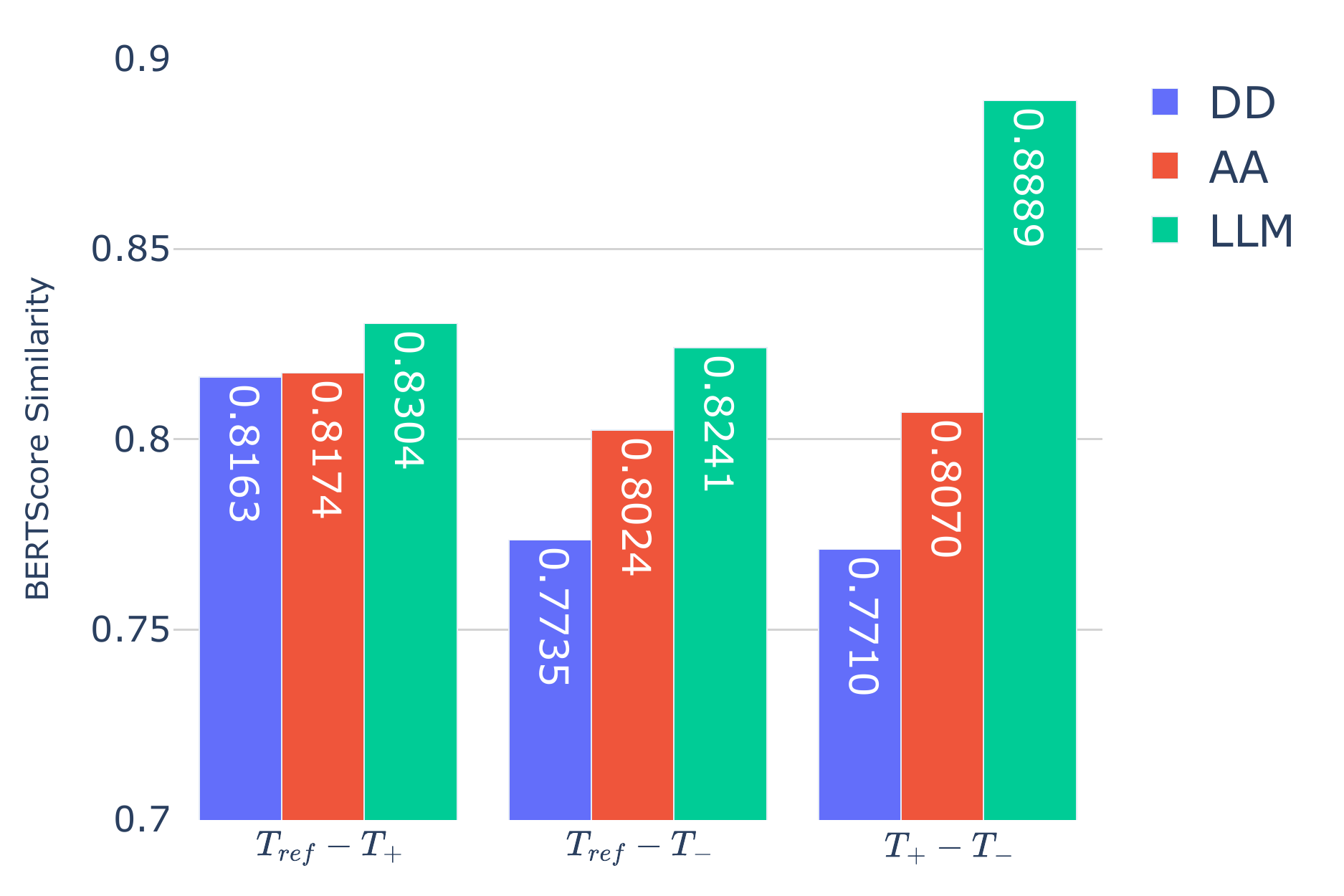}
	\caption{Average pairwise \texttt{BertScore} \cite{zhang2019bertscore} similarity across $T_{ref}$, $T_+$, and $T_-$. Semantic overlap between the candidate set {$T_+$, $T_-$} is significantly higher for the \textit{LLM} compared to other two.} \label{fig:bertscore}
    \vspace{-1em}
\end{figure}

\paragraph{\textit{LLM} as a challenging evaluation setting.}
We attribute the lower performance of evaluation metrics in the $LLM$ evaluation setting to two main factors: (a) both $T_+$ and $T_-$ are generated by the same model for the same query ($q_{ref}$), differing only in the inclusion of reference style text for $T_+$. 
This is perhaps expected, as generation models exhibit inherent stylistic biases \cite{reinhart2025llms}, and hence the outputs may share overlapping stylistic features, complicating discrimination. Also, (b), unlike $DD$ and $AA$, where the input triplet $\{T_{ref}, T_+, T_-\}$ differs in content, all texts in the LLM setting correspond to the same writing task, making style evaluation inherently more difficult. 
As shown in Figure \ref{fig:bertscore}, the \textit{LLM} setting also exhibits a substantial increase in semantic similarity as per its \texttt{BertScore} between $T_+$ and $T_-$. This further highlights the challenge of distinguishing stylistic differences in this scenario.

\begin{figure}[t]
	\centering
	\includegraphics[width=\linewidth]{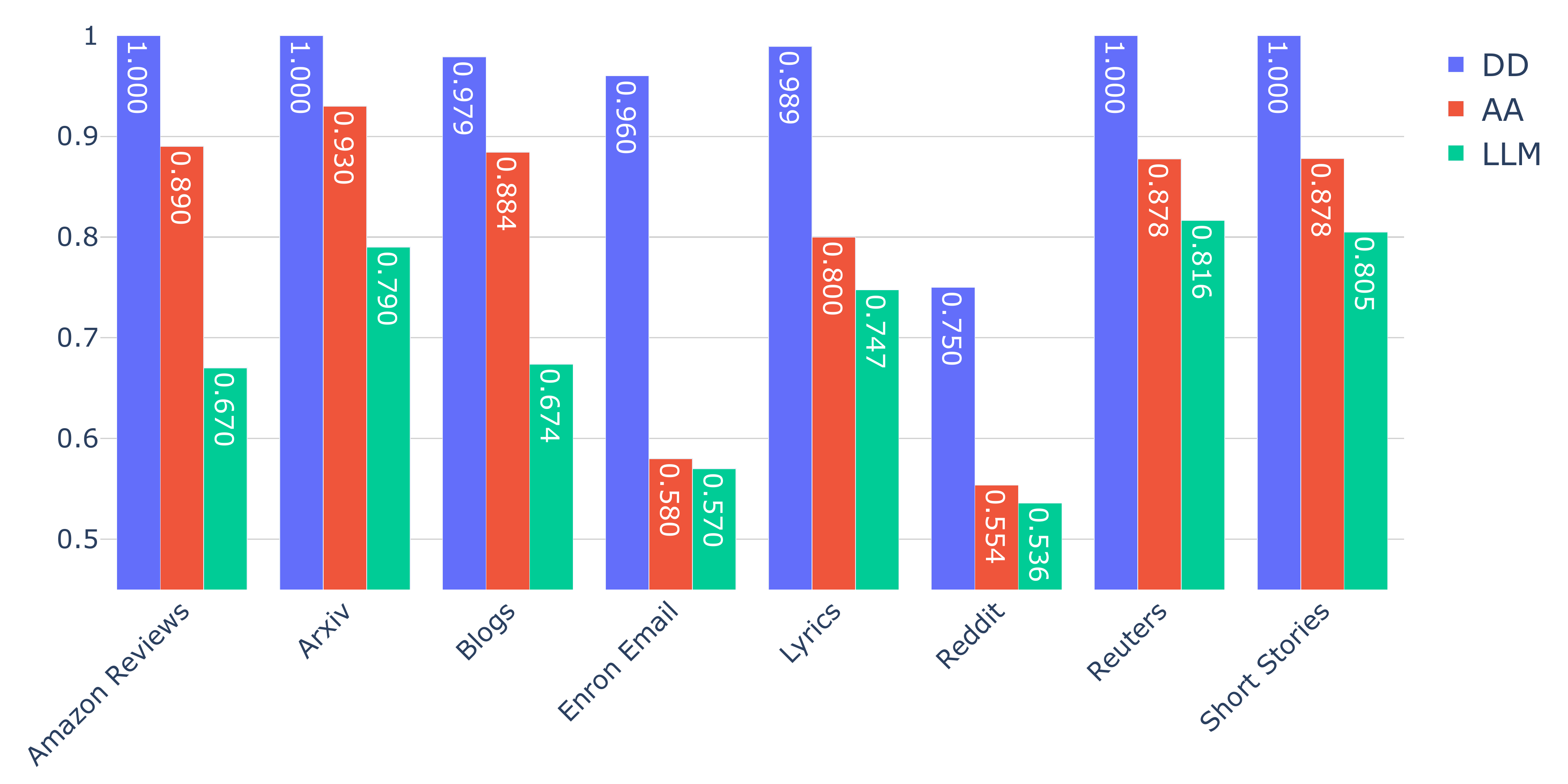}
	\caption{Accuracy of $\rho_{all}(PWV)$ across all domains. \textit{Enron emails} and \textit{Reddit microblogs} achieve lowest accuracy for $AA$ and $LLM$ evaluation settings, marginally outperforming the \texttt{Random} baseline.} \label{fig:domain}
\end{figure}

\paragraph{Analysis of best-performing ensembles}
To assess the contribution of each constituent metric, we computed the agreement between $\rho_{all}(PWV)$ and its individual components: \texttt{BLEU} (0.679), \texttt{ROUGE1} (0.675), \texttt{StyleDistance} (0.664), and \texttt{gpt-4.1} (0.780). 
All metrics contributed approximately equally, and the higher agreement of \texttt{gpt-4.1} can be attributed to the PWV strategy.

After investigating domain-wise accuracy for $\rho_{all}(PWV)$ (Figure \ref{fig:domain}), we observed that the \textit{Reddit} subset yielded the poorest performance, even in $DD$. 
This is likely due to the short length of reference texts (Table \ref{tab:our_data}). 
The accuracy further drops to 0.536 in the \textit{LLM} setting, barely surpassing the \texttt{Random 100} baseline. 
A similar trend is in the \textit{Enron} emails, which also contain short texts and substantial content overlap across authors, reflecting a homogeneous corporate email style. 
In other domains within \textit{LLM}, there was a clear divide between standard writing domains (\textit{arXiv}, \textit{Reuters}, \textit{short stories}) and informal domains (\textit{Amazon food reviews}, \textit{blogs}), with roughly a 0.1 accuracy gap. 
Manual inspection showed that generated responses often failed to capture certain stylistic features, such as ellipses (...), non-standard capitalization, and niche slang (see, e.g., Figure \ref{fig:case_study}).\looseness=-1

\paragraph{Unexpected strength of n-gram metrics.}
Contrary to the evidence from prior work in text style transfer \cite{pang2018unsupervised}, our findings revealed that \textit{n}-gram-based evaluation metrics performed well in the low-resource style discrimination task. We hypothesize that this could be due to the difference in evaluation setup, and attribute this to their evaluation focusing on sentence-level style transfer with limited reference n-grams. This, in turn, prevented the n-gram overlap metrics from capturing broader, more frequent, and nuanced stylistic features within longer spans of text. Our quantitative investigation of different writing domains supports this, as the metrics performed worse for domains with smaller reference texts compared to domains with longer reference texts. 

Particularly for the \textit{LLM} setting, another possible explanation is data contamination. The constituent domain datasets for this work outdate the knowledge cutoff for the generation model $M$ in the \textit{LLM} setting. This could potentially lead to more superficial n-gram overlap between generated and reference texts. We leave further investigation of this possibility using newer datasets to future work.


\paragraph{Explaining performance decline across settings in style embedding-based metrics.} From Table \ref{tab:base_results}, we observe that the performance drop from the $DD$ setting to the $LLM$ setting is the highest for style embedding-based metric (-38.3\%) compared to n-gram (-26.7\%) and LLM-as-judge metrics (-11.3\%). Both style embedding frameworks, \texttt{Wegmann} and \texttt{StyleDistance}, are trained with a contrastive learning objective to separate texts with varying style. This training paradigm leads to a significantly higher performance in the $DD$ task. However, since the writing query for both candidates in the $LLM$ setting is the same, these metrics fail to distinguish between the reference and candidate texts. This is further evident when we look at the statistics of the cosine similarity across the three evaluation settings. The mean and standard deviation for each evaluation setting are: $DD:$ \texttt{Wegmann}$(\mu=0.69, \sigma=0.47)$, \texttt{StyleDistance}$(\mu=0.19, \sigma=0.13)$, $AA:$ \texttt{Wegmann}$(\mu=0.19, \sigma=0.45)$, \texttt{StyleDistance}$(\mu=0.05, \sigma=0.12)$, and $LLM:$ \texttt{Wegmann}$(\mu=0.05, \sigma=0.25)$, \texttt{StyleDistance}$(\mu=0.01, \sigma=0.05)$. We observe that the difference $Sim_{cos}(\vec{T_{ref}}, \vec{T_+}) - Sim_{cos}(\vec{T_{ref}}, \vec{T_-})$ is the highest for the $DD$ setting, then $AA$, and then $LLM$; indicating that the style embeddings are least confident in their style discrimination predictions in $LLM$.

\paragraph{Limitations of LLM-as-judge} 
Our analysis of different prompting strategies to improve LLM-as-judges yielded positive, yet inconsistent, outcomes.
Positive outcomes were due to APO boosting accuracy by about +25\%. 
However, remark that this, as an improvement, could be a false positive: the performance for in \textit{AA} for the $P_{struct}$ and $P_{binary}$ prompts are 0.238 and 0.491, respectively, providing close to random performance in their output space. 
Hence, APO on these prompts also does not improve accuracy significantly, if at all. 
This is expected, since adaptive prompting strategies like APO have been shown to be prone to overfitting to the training instances \cite{de2025context}. 
Limiting the model output to a binary choice, instead of allowing responses like ``Both'' and ``None'', yielded about a +35\% increase, although it did not improve the model's reasoning capability in the style discrimination task. 
It then follows that LLMs-as-judges are, as it is well-known, difficult to reliably apply to this task.

\paragraph{IAA Human Analysis}
The agreement amongst humans revealed that style preferences are more subjective than content preferences, which aligns with previous work \cite{salemi2025expert}. Likewise, style-personalized preferences vary across individuals and per person. 
This can be seen in the low agreement found in copy-editing when compared to that of the other two preference tasks.


\section{Conclusion}

SPTG evaluation in the LLM era is challenging due to the dependency on pragmatics and the complexities associated with non-standardized evaluation metrics, intricate annotation requirements, and data scarcity.

Our experiments revealed several limitations in both widely adopted and newly investigated evaluation paradigms. 
Namely, \textbf{individual metrics did not have good agreement} in settings: although they were most effective at distinguishing texts across domains, they were only moderately effective at within-domain authorship attribution, and least effective at distinguishing personalized and non-personalized LLM-generated text. 
However, \textbf{ensembling metrics} matched or outperformed their individual constituents. 

Based on our findings, we argue that more careful scrutiny is required when evaluating SPTG. 
This is reinforced by our findings in the human IAA ablation study: due to pragmatic dependencies, \textbf{the complexity of measuring SPTG is exacerbated by non-standardized, low-agreement metrics}. 
It then follows that emphasis should be on developing reliable metrics that better adjust to the \textit{problem}, rather than the dataset evaluated, in order to properly measure SPTG and make research comparable. 
\section*{Limitations}

While we introduce a comprehensive evaluation benchmark for assessing style-personalized text generation, our work has three primary limitations. \textbf{First}, we focus exclusively on the binary classification task of discriminating between two candidate responses given a reference text, leaving the development of fine-grained and explainable evaluation systems to future work. We argued, however, that our focus was on correlation between metrics, and a more fine-grained measure would make agreement more complex and less reliable. 
\textbf{Second}, we only explore two ensembling approaches: majority voting and performance-weighted voting, leaving exploration of other ensembling techniques like likelihood-based voting, threshold-tuned voting and meta-ensemble learners to future work. Our findings did support that simple ensembling techniques performed better than other metrics, and thus we argue that more effective ensembles can only benefit SPTG measurement. 
\textbf{Third}, the binary task formulation does not fully capture the first-person perspective inherent to style-personalized text generation. It is difficult to build a diverse, user-specific first-person evaluation benchmark; especially when it is necessary that it be out-of-domain to counteract any potential memorization by LLMs. Instead, however, we adopted a set of simpler style discrimination tasks to evaluate both existing and proposed metrics. 

\section*{Ethics}
Our work is an analysis of the measurements for SPTG. 
We are unaware of any potential misuse of our work, although we recognize that we could not have covered all possible scenarios. 
All the datasets we collected were licensed under \href{https://creativecommons.org/licenses/by/4.0/}{CC BY 4.0}, \href{https://creativecommons.org/licenses/by-sa/4.0/deed.en}{CC BY-SA 4.0}, \href{https://creativecommons.org/publicdomain/zero/1.0/}{CC0}, and \href{https://www.apache.org/licenses/LICENSE-2.0}{Apache Version 2.0}. We do not release any content from these datasets, and will only release a reproducibility script for future research work in the camera-ready version. For our human annotation study, we recruited participants through a professional crowdsourcing platform. All participants were over 18 years old, and prior to the annotation study their consent was collected electronically. The annotators were compensated at an hourly rate of 12 USD and were advised to take breaks as needed to maintain focus and comfort throughout the task. This study was approved by the institutional review board of the authors' organization(s).


\DeclareRobustCommand{\DE}[3]{#2}
\bibliography{main}

\appendix
\section{Call Parameters}\label{app:callparameters}

For our meta-prompting approach we used \texttt{gpt4o} for both $f_{query}$ and $M$, with temperature set to one. The calls were made in August 2024. 
For our LLMs-as-judges, we used \texttt{Ministral-3B}, \texttt{Llama-3.1-8B} \cite{grattafiori2024llama}, \texttt{Mistral-Small-24B-Instruct-2501}, \texttt{Qwen3-32B} \cite{yang2025qwen3}, \texttt{DeepSeek-V3-0324} \cite{liu2024deepseek}, \texttt{o4-mini} and \texttt{gpt-4.1} \cite{hurst2024gpt}. 
For classification, we set the temperature to zero whenever possible; and the maximum output token length to 16192 tokens or their maximum allowable\footnote{We set a high maximum output token length to accommodate thinking model responses, and used the same value for non-thinking models for comparable response.}.
For APO, we used the implementation by \cite{pryzant2023automatic} with beam size 4, training batch size 64, search depth 6 and keep the rest of the hyperparameters the same as the original paper. All temperatures in the algorithm were set to 0.7, except during the classification phase, where we set it to zero. 
We use the same training and test sets across all experiments. 
\section{LLM-as-a-Judge Responses}\label{app:llmsresponses}

\begin{figure*}[h]
	\centering
	\includegraphics[width=0.7\linewidth]{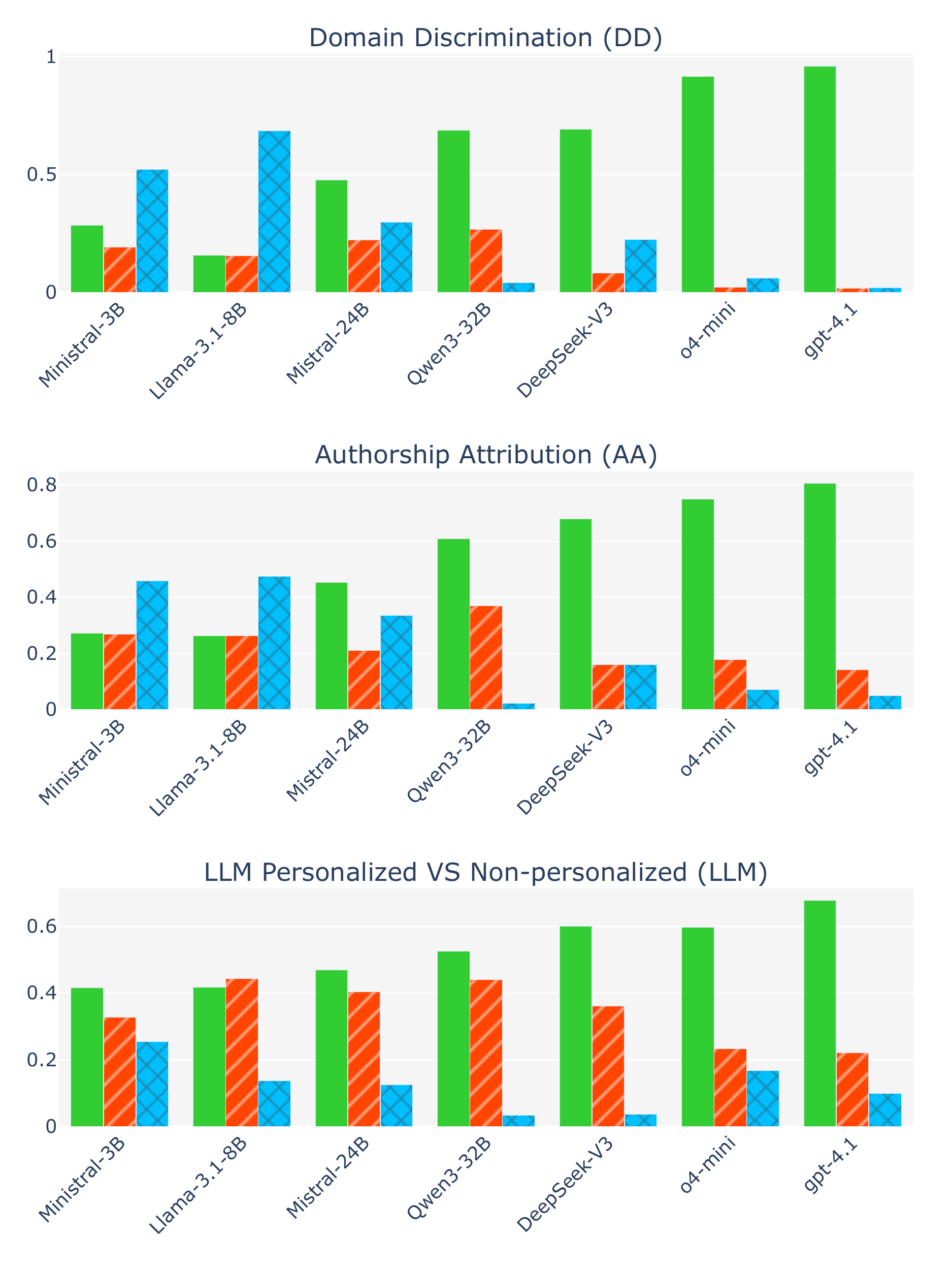}
	\caption{Distribution of LLM-as-judge evaluation metric responses across different evaluation settings. Correct classification is denoted by solid \textcolor{green}{green} bars, misclassification is denoted by diagonal \textcolor{red}{red} bars, and indifferent by cross-hatch \textcolor{blue}{blue} bars. A response is marked ``Indifferent" if it does not respond with $T_+$ or $T_+$, including other responses like ``Both'' and ``None'' (see Figure \ref{fig:eval_prompt}).}
	\label{fig:llm_as_judge}
\end{figure*}

The class distribution (evaluation metric responses) for LLMs-as-judges are in Figure~\ref{fig:llm_as_judge}. It can be seen that the model performance increases with increase in number of parameters consistently across all three evaluation settings. Smaller models like \texttt{Ministral-3b} and \texttt{Llama-3.1-8b} generate a higher degree of indifferent responses (including responses like ``Both'' and ``None'') compared to larger models. We can also note that the misclassification error rate increases from the $DD$ setting to the $AA$ setting, and is the highest for the $LLM$ setting.

\begin{figure*}[t]
    \centering
    \begin{subfigure}{0.75\linewidth}
        \centering
        \includegraphics[width=\linewidth]{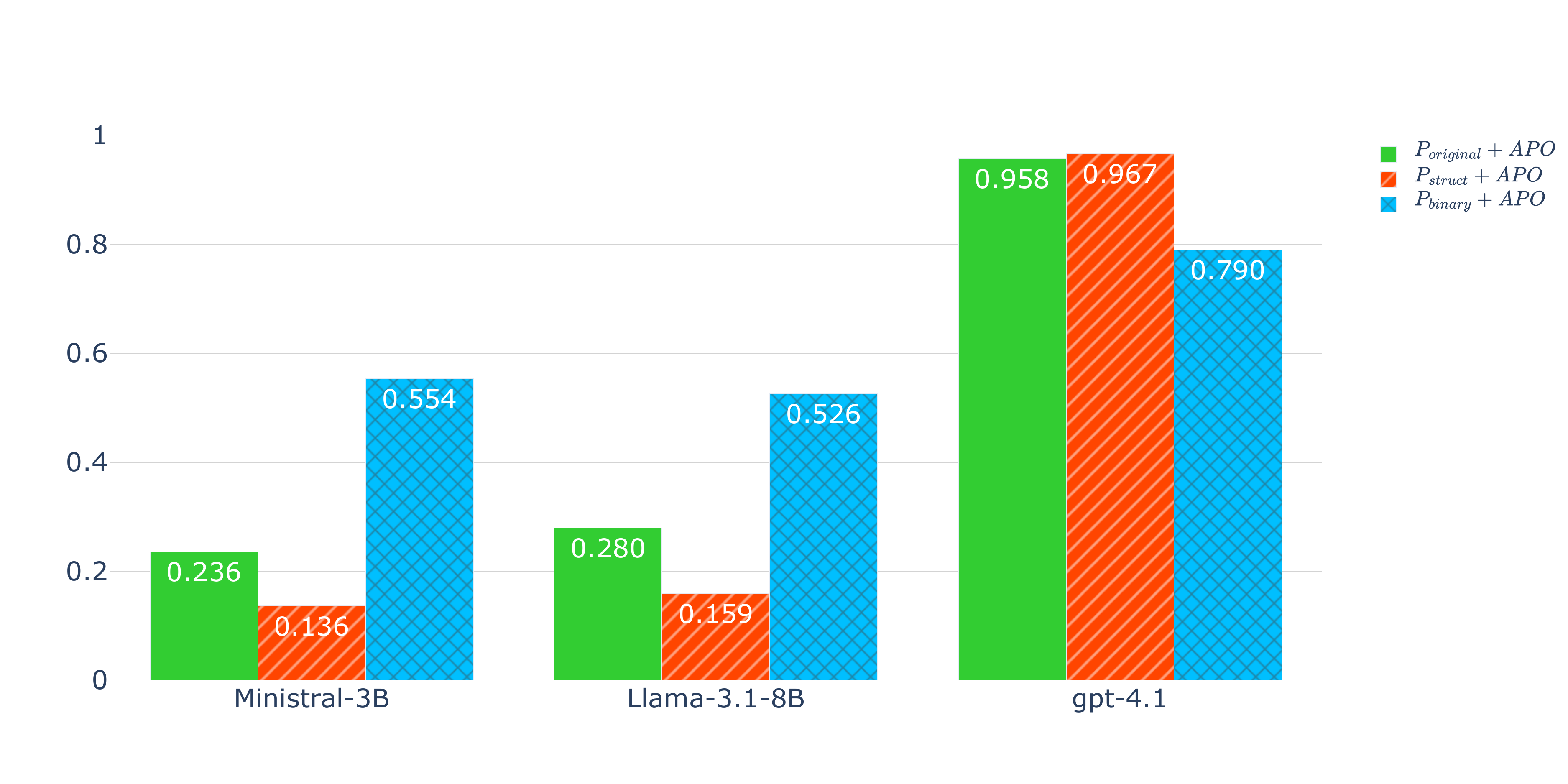}
    \end{subfigure} \\
    \begin{subfigure}{0.8\linewidth}
        \centering
        \includegraphics[width=\linewidth]{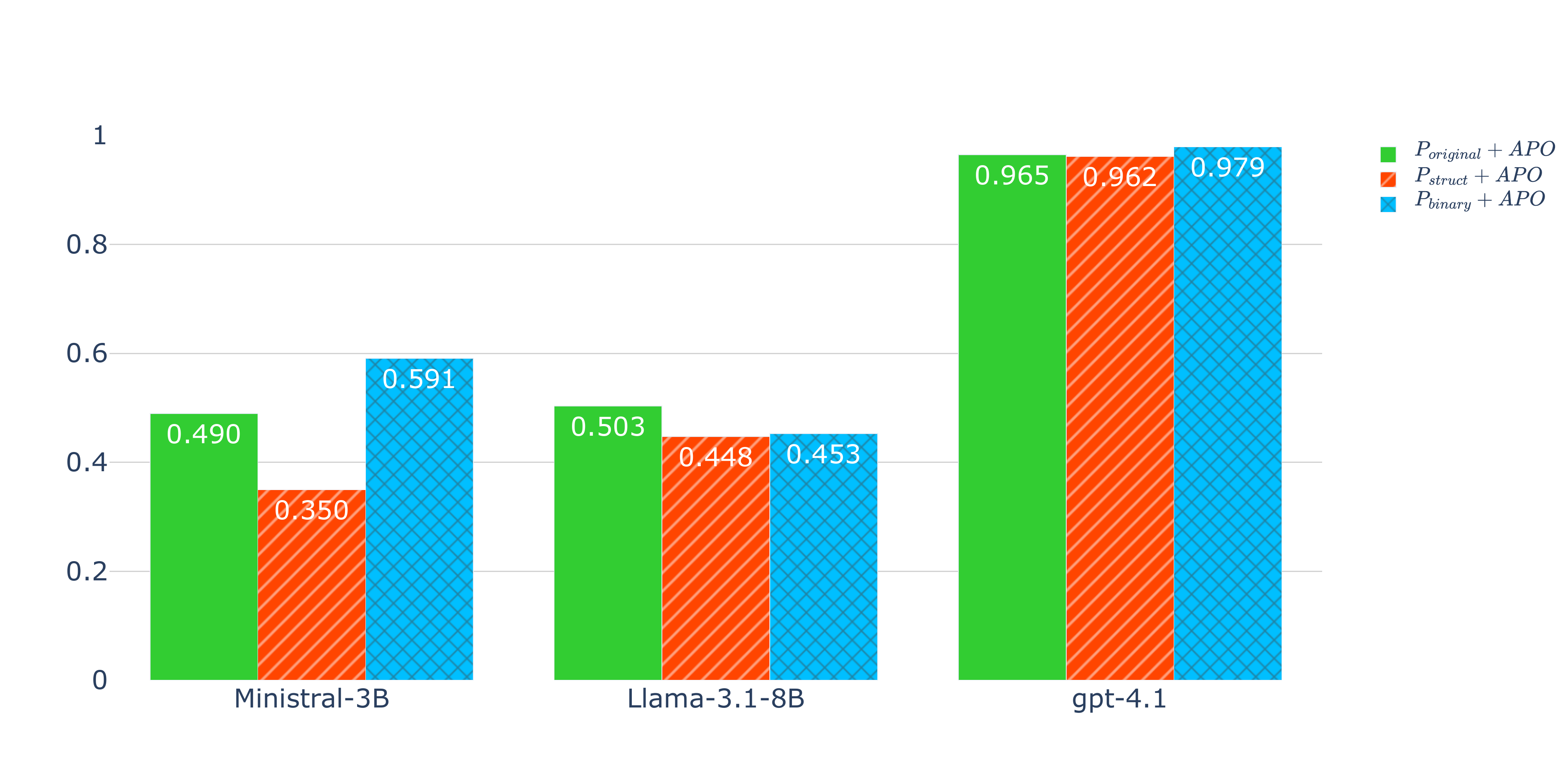}
    \end{subfigure}
    
    \caption{Ablation study results for LLM-as-judges prompting for the \textit{domain discrimination (DD)} evaluation setting for \texttt{Ministral-3B}, \texttt{Llama-3.1-8b}, and \texttt{gpt-4.1} models. Top: LLM-as-judge results for different prompting strategies $P_{orig}$ (solid \textcolor{green}{green} bars), $P_{struct}$ (diagonal \textcolor{red}{red} bars), and $P_{binary}$ (cross-hatch \textcolor{blue}{blue} bars). Bottom: automatic prompt optimization LLM-as-judge results.}
    \label{fig:human_eval}
\end{figure*}
\section{Prompts}\label{app:prompts}

\subsection{Data Generation}
\begin{figure*}[t]
	\centering
	\includegraphics[width=0.7\linewidth]{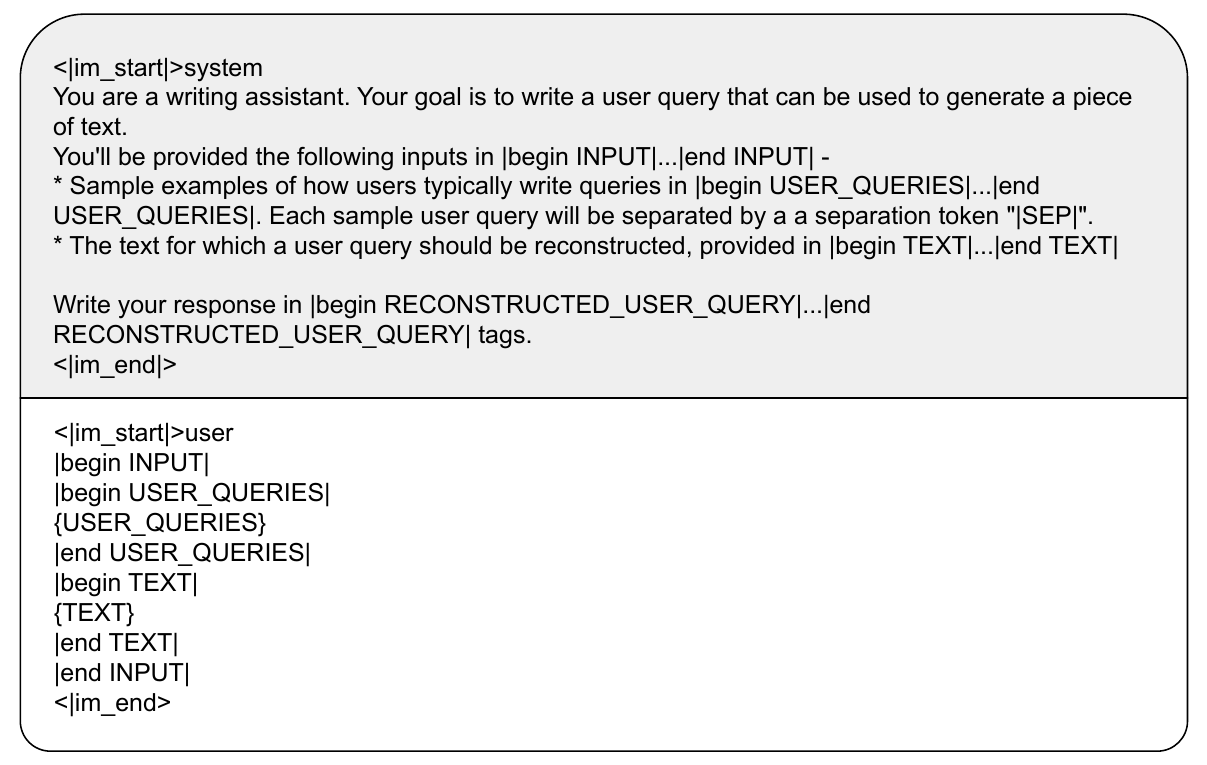}
	\caption{Query reconstruction prompt in chat markup format.}
	\label{fig:query_reconstruction_prompt}
\end{figure*}

\begin{figure*}[t]
	\centering
	\includegraphics[width=0.9\linewidth]{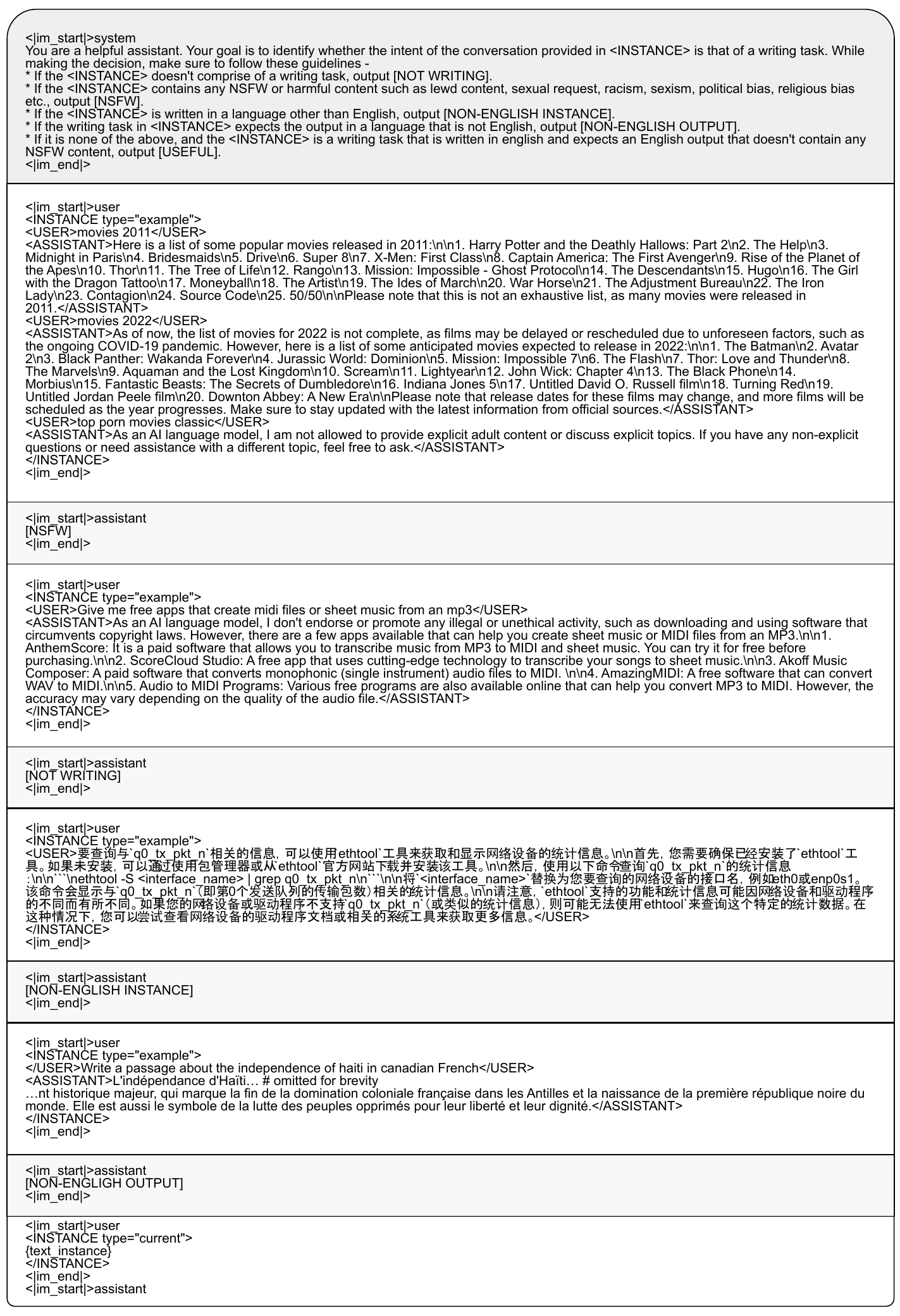}
	\caption{Classification prompt used to filter out English writing queries from the WildChat dataset \cite{zhao2024wildchat}.}
	\label{fig:is_writing}
\end{figure*}

\begin{figure*}[t]
	\centering
	\includegraphics[width=0.9\linewidth]{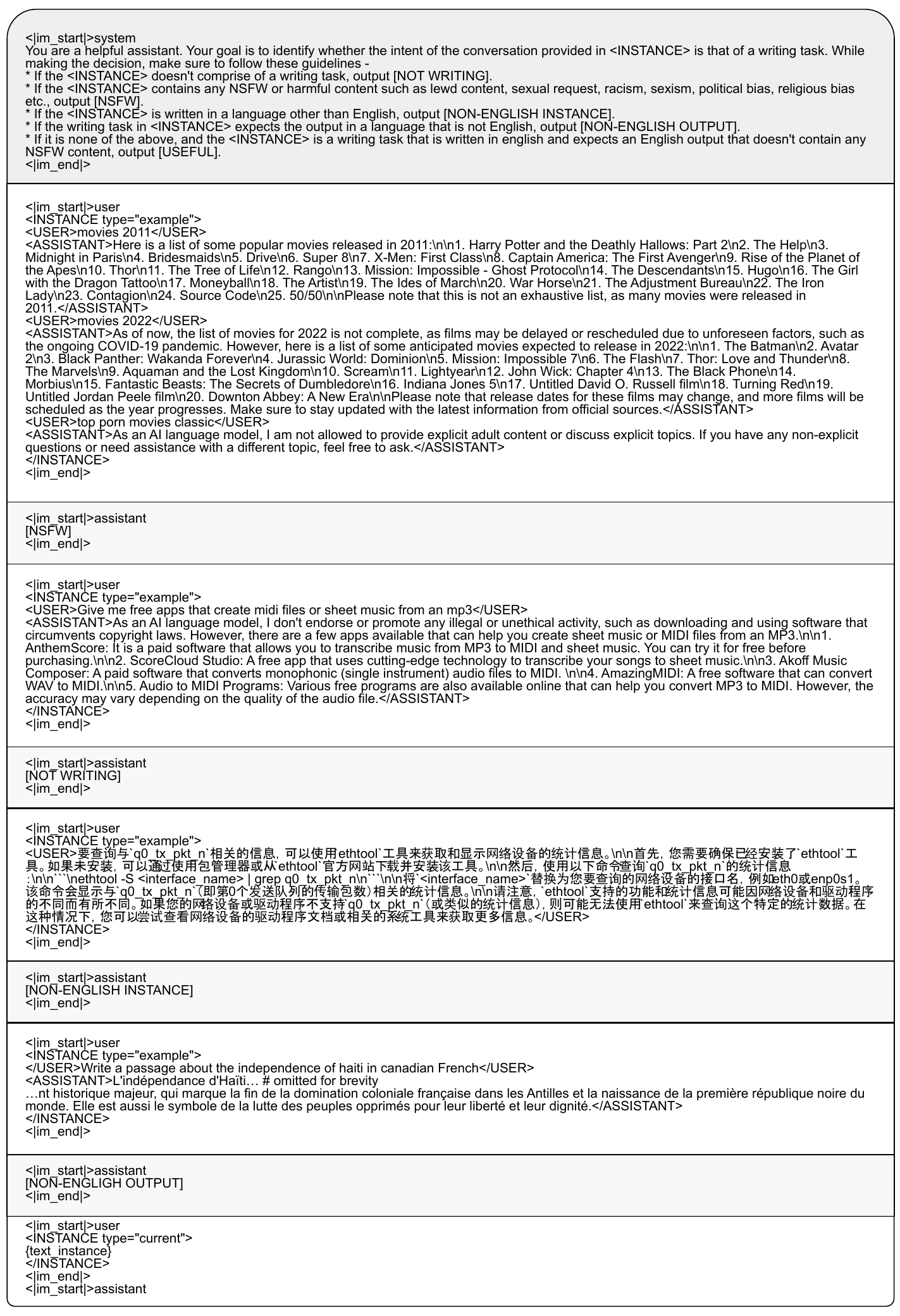}
	\caption{Classification prompt used to obtain the writing task corresponding to the filtered English writing queries from the WildChat dataset \cite{zhao2024wildchat}.}
	\label{fig:task_classification}
\end{figure*}

The prompts for filtering are in Figures \ref{fig:is_writing} and \ref{fig:task_classification}). 
The prompt to generate the reconstructed user query $q_{ref}$ is in Figure \ref{fig:query_reconstruction_prompt}). 
We provide the respective outcomes of filtration steps on the subset of \texttt{WildChat} user queries in Figures \ref{fig:wildchat_filteration1} and \ref{fig:wildchat_filteration2}.

\begin{figure*}[t]
	\centering
	\includegraphics[width=0.9\linewidth]{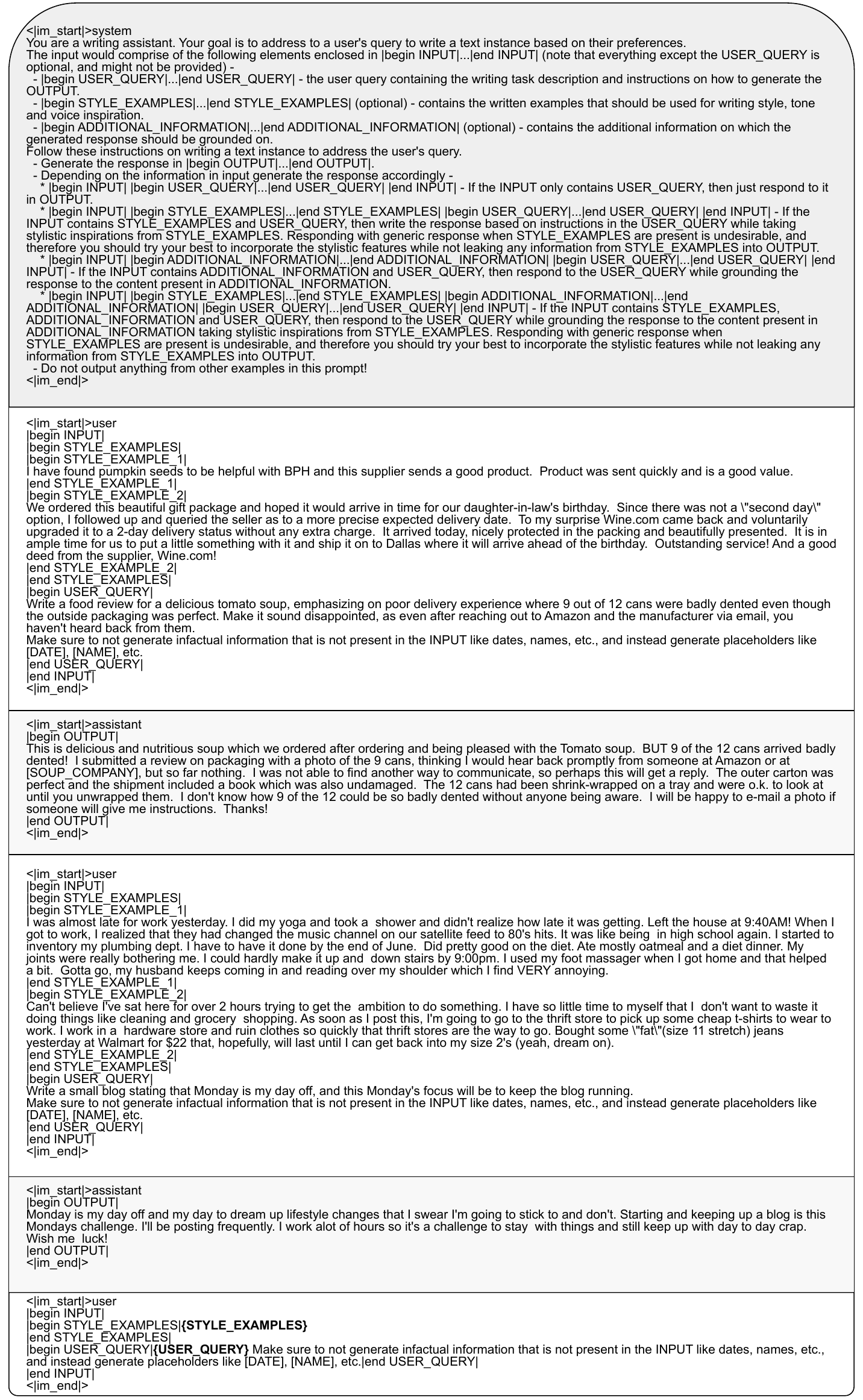}
	\caption{Personalized text generation prompt used to generate $T_+$ for \textit{LLM} evaluation setting in chat markup format.}
	\label{fig:personalized_prompt}
\end{figure*}

\begin{figure*}[t]
	\centering
	\includegraphics[width=0.9\linewidth]{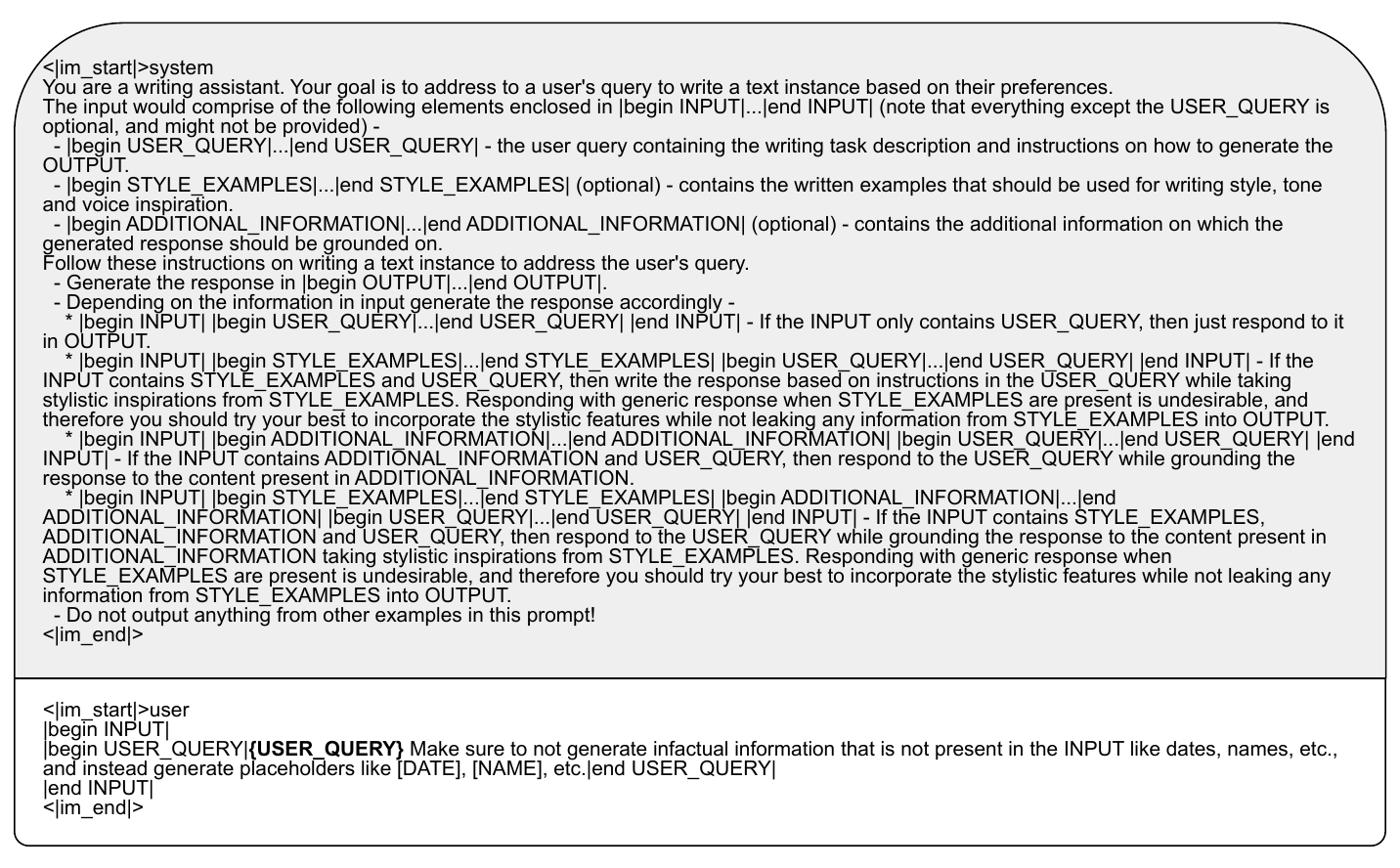}
	\caption{Non-personalized text generation prompt used to generate $T_-$ for \textit{LLM} evaluation setting in chat markup format.}
	\label{fig:nonpersonalized_prompt}
\end{figure*}

Finally, prompts for the texts to be annotated are in Figures \ref{fig:personalized_prompt} and \ref{fig:nonpersonalized_prompt} for the prompts.

\subsection{Evaluation}
\begin{figure*}[h]
	\centering
	\includegraphics[width=0.7\linewidth]{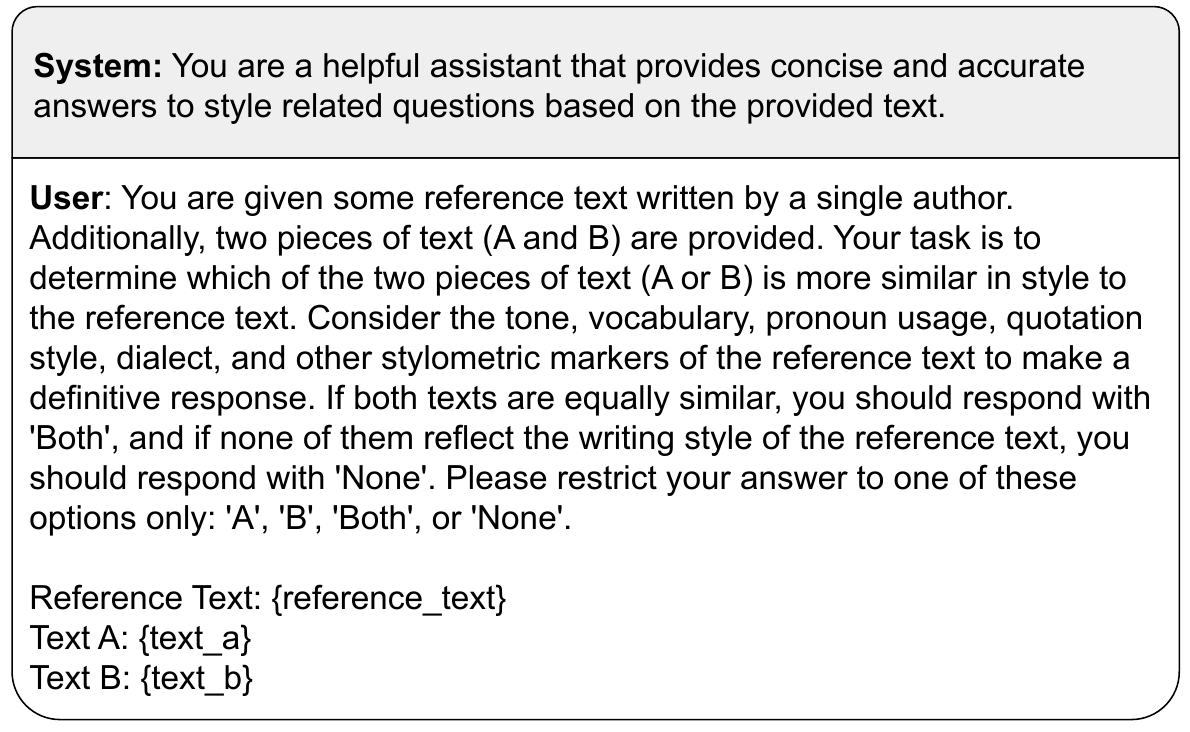}
	\caption{Zero-shot evaluation prompt used to evaluate all LLM-as-judge metrics (Section \ref{subsec:baselines}). We refer to this prompt as $P_{orig}$ in Section \ref{subsec:apo}.}
	\label{fig:eval_prompt}
\end{figure*}

\begin{figure*}[h]
	\centering
	\includegraphics[width=0.65\linewidth]{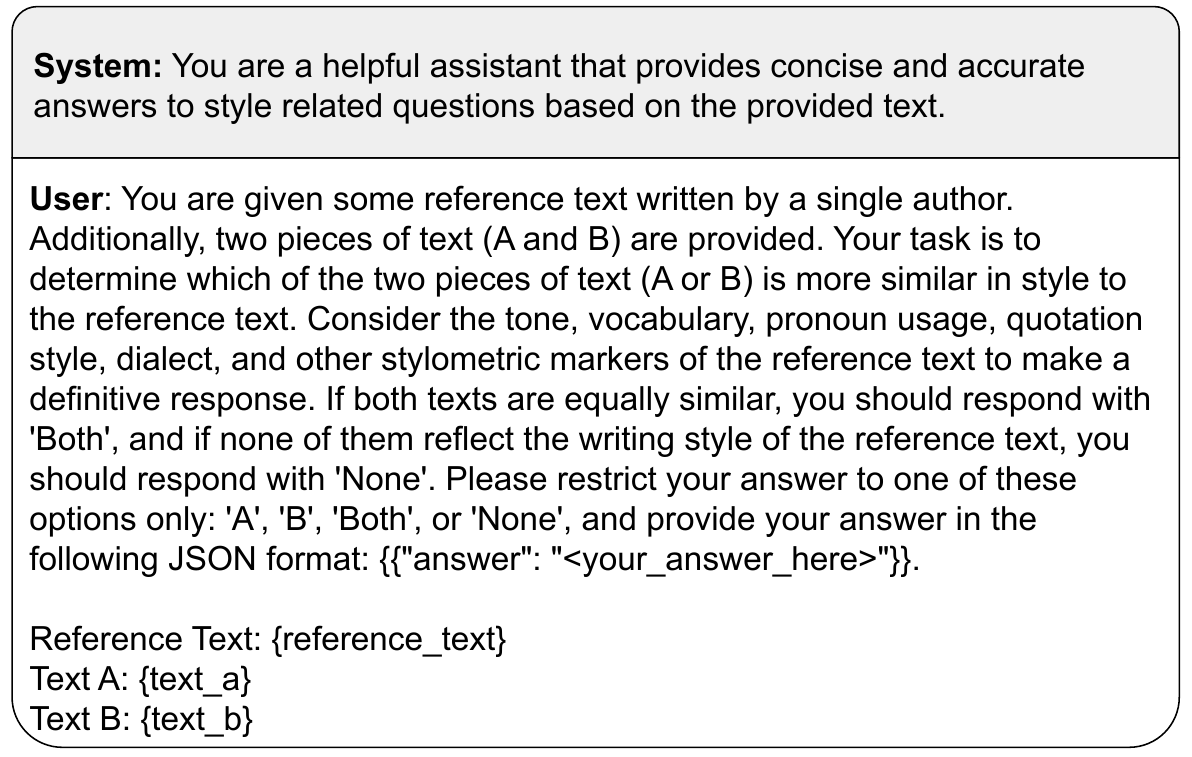}
	\caption{Zero-shot evaluation prompt ($P_{struct}$) used to evaluate parseable structured output in Section \ref{subsec:apo}.}
	\label{fig:eval_prompt_struct}
\end{figure*}

\begin{figure*}[h]
	\centering
	\includegraphics[width=0.65\linewidth]{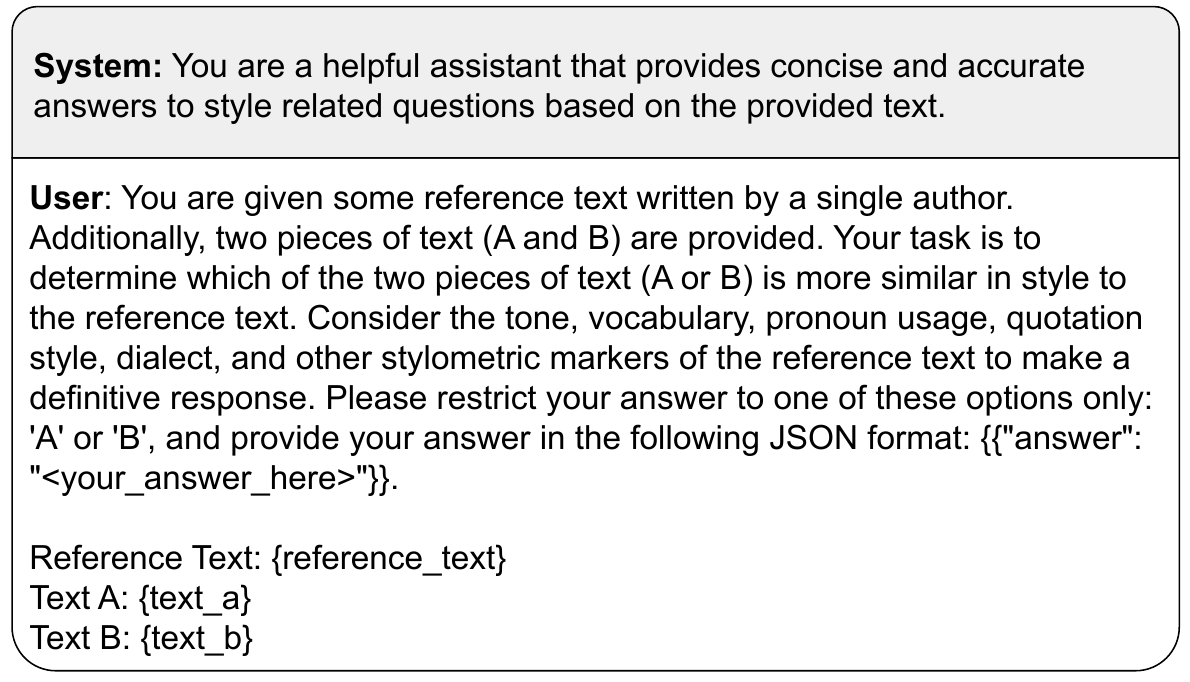}
	\caption{Zero-shot evaluation prompt ($P_{binary}$) used to evaluate binary output in Section \ref{subsec:apo}.}
	\label{fig:eval_prompt_binary}
\end{figure*}

The prompt used for judging is in Figure~\ref{fig:eval_prompt}. For exploration of different prompting strategies (Section~\ref{subsec:apo}), we use the prompts in Figures~\ref{fig:eval_prompt},~\ref{fig:eval_prompt_struct}, and~\ref{fig:eval_prompt_binary} as $P_{orig}$, $P_{struct}$ and $P_{binary}$ respectively. 

\section{Dataset Statistics}

\begin{figure*}[t]
	\centering
	\includegraphics[width=0.7\linewidth]{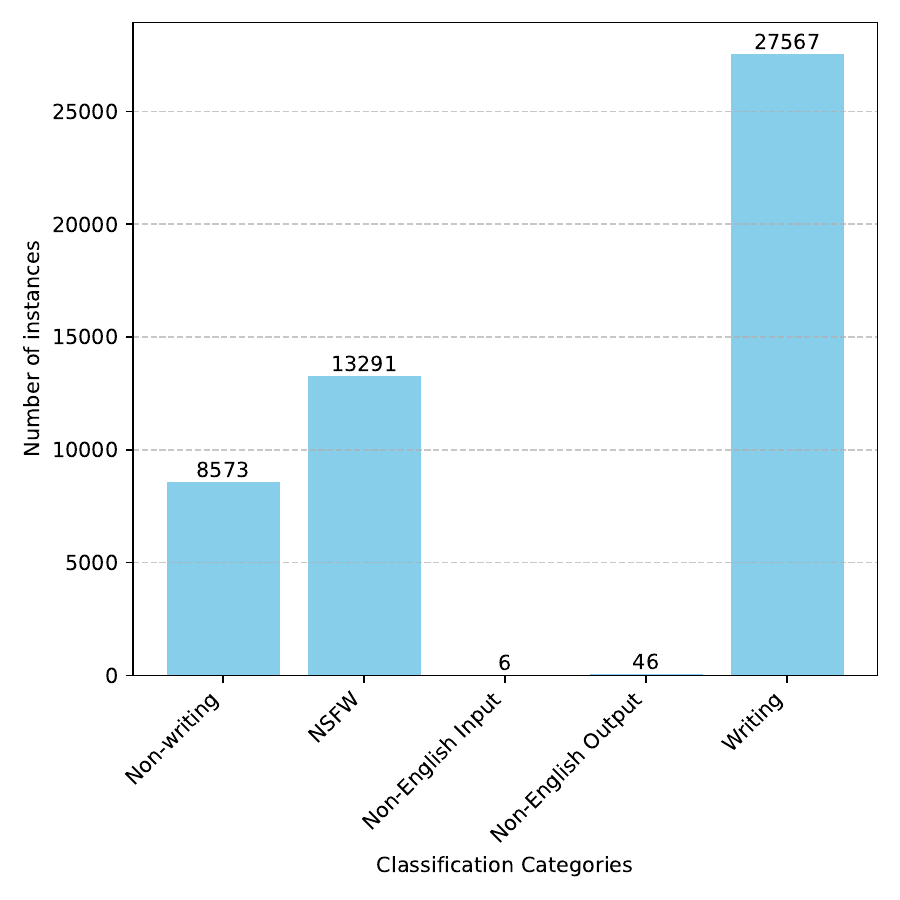}
	\caption{WildChat user query classification statistics for writing task identification over randomly selected 50,000 instances.}
	\label{fig:wildchat_filteration1}
\end{figure*}

\begin{figure*}[t]
	\centering
	\includegraphics[width=\linewidth]{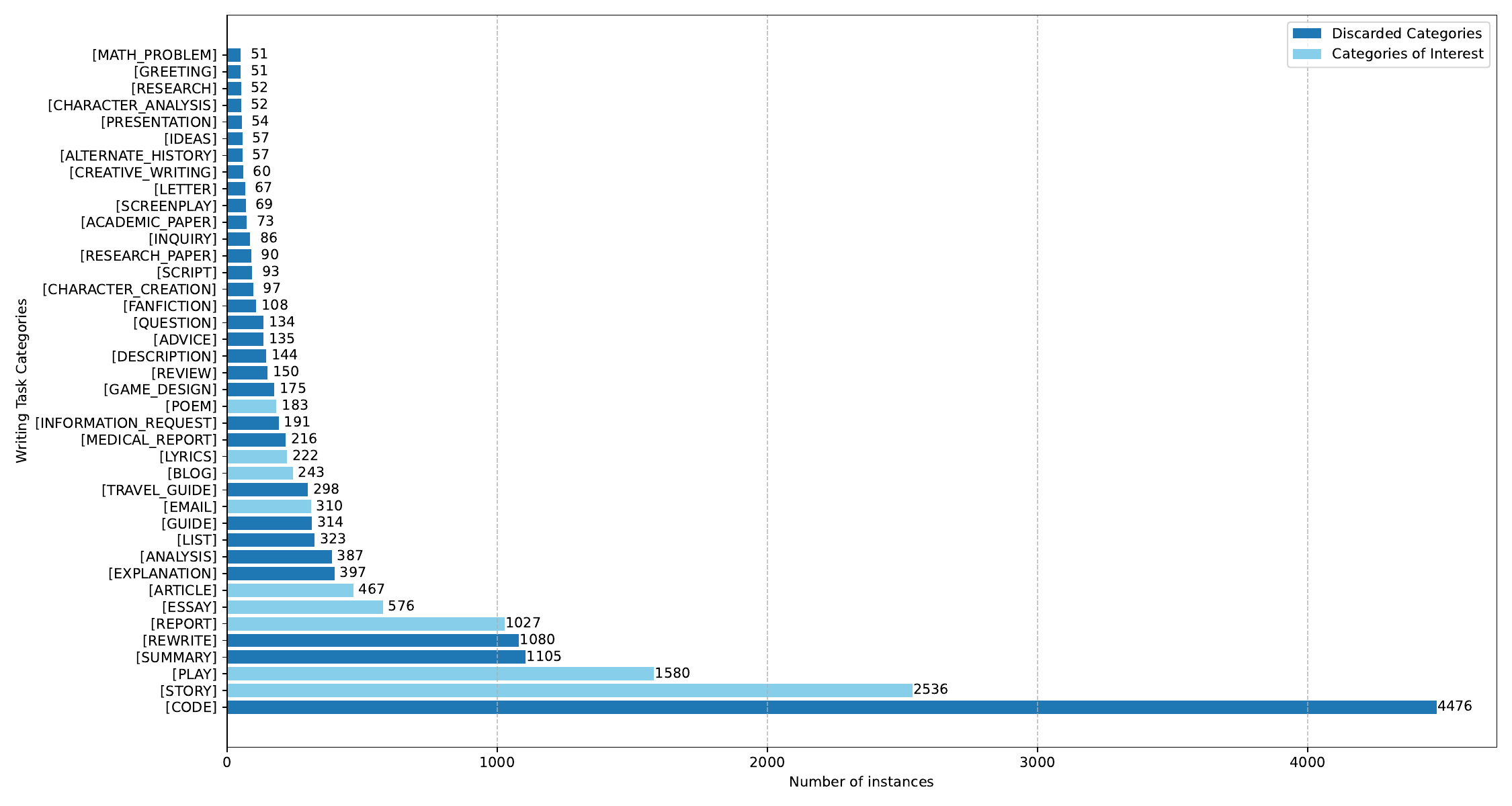}
	\caption{WildChat user query classification statistics for categorizing the writing queries obtained from Figure \ref{fig:wildchat_filteration1}.}
	\label{fig:wildchat_filteration2}
\end{figure*}

The data statistics for our filtered data are in Figures~\ref{fig:wildchat_filteration1} and~\ref{fig:wildchat_filteration2}. From the first filtration step, we filter out 17\% of the non-writing queries, and 26\% of queries marked as not safe for work (NSFW), keeping 27,567 queries for the second filtration step. We select the top 40 most frequent labels from the second filtration step, and retain the queries that resemble the writing tasks of our style discrimination dataset. When selecting seed user queries for the query reconstruction function, $f_{query}$, we map \textit{Amazon reviews} and \textit{arXiv abstracts} domains to [REPORT] queries, \textit{blog} domain to [BLOG] queries, \textit{Enron emails} domain to [EMAIL] queries, \textit{lyrics} domain to [LYRICS] queries, \textit{short stories} domain to [STORY] and [PLAY], \textit{Reddit microblogs} domain to [ESSAY] queries, and \textit{Reuters news articles} to [ARTICLE] and [REPORT] queries. 

\begin{figure*}[h]
	\centering
	\includegraphics[width=0.9\linewidth]{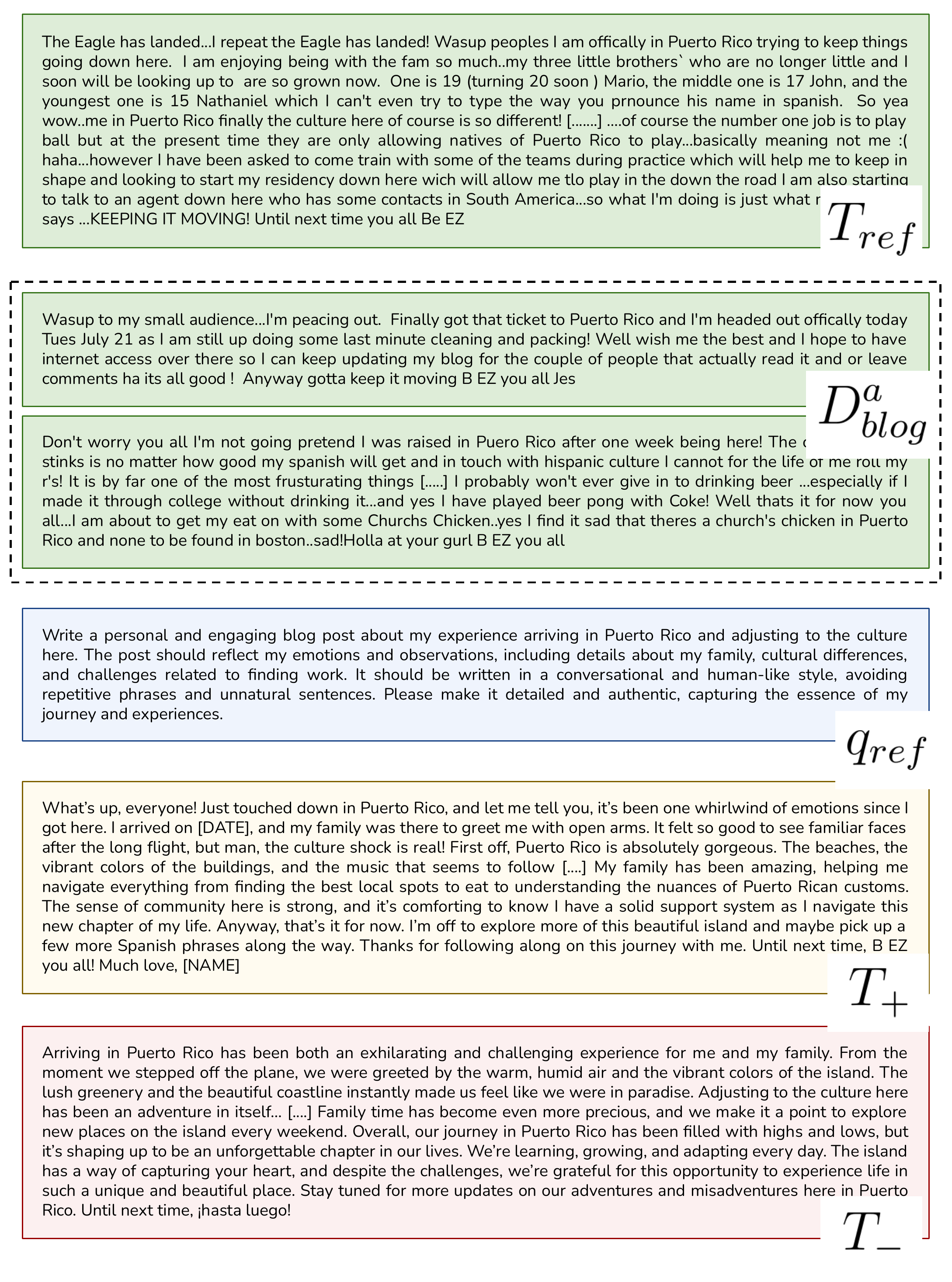}
	\caption{Example instance of our evaluation dataset from the \textit{blogs} domain. The candidates $T_{+}$ and $T_{-}$ are generated for the \textit{LLM} evaluation setting, with $T_{+}$ denoting the personalized response and $T_{-}$ denoting the non-personalized response.}
	\label{fig:case_study}
\end{figure*}

We provide an example from our dataset for the \textit{blog} domain in Figure \ref{fig:case_study}, as well as the licensing details of the source datasets in Table \ref{tab:data_license}.

\begin{table}[]
\caption{\revision{Licensing status of individual data source.}}\label{tab:data_license}
\centering
\begin{tabular}{lc}
\toprule
\textbf{Dataset Name} & \textbf{License} \\
\midrule
\textbf{\textit{Amazon}}  & \href{https://creativecommons.org/publicdomain/zero/1.0/}{CC0: Public Domain} \\
\textbf{\textit{ArXiV}} & \href{https://creativecommons.org/publicdomain/zero/1.0/}{CC0: Public Domain} \\
\textbf{\textit{Blogs}} & Unspecified\footnote{} \\
\textbf{\textit{Enron}}  & \href{https://creativecommons.org/licenses/by/3.0/us/}{CC-BY-3} \\
\textbf{\textit{Lyrics}}  & \href{https://creativecommons.org/licenses/by-sa/4.0/}{CC BY-SA 4.0} \\
\textbf{\textit{Reddit}} & \href{https://www.apache.org/licenses/LICENSE-2.0}{Apache License, Version 2.0} \\
\textbf{\textit{Reuters}} & \href{https://creativecommons.org/licenses/by/4.0/}{CC BY 4.0} \\
\textbf{\textit{Short Stories}} & \href{https://creativecommons.org/licenses/by/4.0/}{CC BY 4.0} \\
\bottomrule
\end{tabular}%
\end{table}

\footnotetext[7]{The Blog Authorship Corpus 
\href{https://u.cs.biu.ac.il/~koppel/BlogCorpus.htm}{website} states, 
"\textit{The corpus may be freely used for non-commercial research purposes}".}

\section{Challenges in Collecting Human Annotations for Long-Form SPTG}\label{app:humanannotation}

As mentioned in passing, SPTG is difficult to measure, in part due to the scarcity of data as well as the \textit{personal} nature of the task. 
While the main body of our work focused on \textit{automated} measures, we also conducted a brief two-part, side-by-side study on human preferences. 
On the first part, we measured human preference on \textit{content preference} (which generated content better responded to the prompt); \textit{style preference} (which content better adjusted to the style of the grounding text); and \textit{copy-editing} (which of the outputs would be most useful). 
The second part was identical to the first, including the same annotators, although performed a month after the study. 

We found that, as measured by the agreement ratio, although human annotators had high agreement on content preference (0.779 IAA) and moderate-to-high on style preference (r. 0.641), their agreement on copy-editing preference was low (r. 0.400). Remark that it is \textit{comparatively} low, since the label set is a five-label set and thus chance agreement would be 0.2. The second study reproduced the same ordering, albeit with slightly different numbers (0.736, 0.615, and 0.385 respectively). 

This revealed two key findings: first, given that content preference was higher than style preference (and more reproducible), \textbf{style preferences are more subjective than content preferences}. This aligns with previous work on SPTG evaluation \cite{salemi2025expert}. 
Second, \textbf{style-personalized preferences vary across individuals and per-person}. 
This can be seen in the low agreement in copy-editing in both rounds. 
Both takeaways highlight one of the core difficulties of long-form SPTG, and reinforce our emphasis on the need for better, standardized measurements: human annotation, \textit{especially when the annotators are not the authors}, is unpredictable. 
Having non-standardized metrics with low agreement amongst each other only makes the problem--and the comparability of works--more difficult.

Next we describe our study in detail. 

\paragraph{First Part: Corpus Annotation}
We collected labels for SPTG emphasizing three key aspects: content preference, style preference, and copy-editing preference. 
For each labeling task, the annotators were given a reference text for a given author's writing style (denoted ``\textit{|Grounding Text|}'' in the example below), a user query (r. ``\textit{|User Input|}''), and two generated responses based on the query: one personalized and another non-personalized.  
To avoid positional bias, each generated output was randomly marked as ``\textit{|Generated Output A|}'' or ``\textit{|Generated Output B|}''.

The task involved answering each of the three key aspects for our labeling task. They were asked, in order, for content preference

\begin{quote} \small
    \textit{Which generated output is better able to cover the content being asked by the |User Input|? \\
    (a) |Generated Output A| \\
    (b) |Generated Output B| \\
    (c) Both generated responses have similar content \\
    (d) None of the generated responses are able to generate a useful response for the |User Input|},
\end{quote}

\noindent for style preference 
\begin{quote} \small
    \textit{Which generated output is better able to reflect the writing style of |Grounding Text|? \\
    (a) |Generated Output A| \\
    (b) |Generated Output B| \\
    (c) Both generated responses have similar writing style to |Grounding Text| \\
    (d) None of the generated responses reflect the writing style to |Grounding Text| \\
    (e) I cannot make out the difference in the writing styles },
\end{quote}

\noindent and, finally, for copy-editing preference
\begin{quote} \small
    \textit{If you were asked to write a response towards |User Input|, which generated output(s) would you use to develop an answer? \\
    (a) |Generated Output A| \\
    (b) |Generated Output B| \\
    (c) I will use a combination of both of them, taking inspirations for certain aspects from each generated response. \\
    (d) I will use either of the two generated outputs. \\
    (e) I will not use any of the generated responses and rather write a response from scratch.}
\end{quote}

Our corpus was comprised of 396 instances, spanning 75 instances from \textit{Amazon reviews}, 71 instances from \textit{blogs}, 68 instances from \textit{lyrics}, 42 instances from \textit{Reddit microblogs}, 35 instances from \textit{Reuters news articles}, and 30 instances from \textit{short stories}\footnote{\textit{ArXiv scientific abstracts} were excluded from our annotation study due to the domain-specific expertise required.}. 
Each instance was labeled by three annotators. 

The IAA scores for the content preference was 0.779, for style preference was 0.641, and for copy-editing preference was 0.400. We provide the distribution of response choices in Figure \ref{fig:human_eval}. 
We found that the annotators had difficulty distinguishing between the personalized and non-personalized responses, often labeling ``Both'' as the most prominent response for all three annotation dimensions. 
We explain this in two ways: (1) there is high content overlap between the personalized and non-personalized generations (recall the high semantic overlap in the \textit{LLM} setting from Figure \ref{fig:bertscore}); 
and (2) annotating long-form text was difficult for the annotators. 
Obtaining high-quality long-form annotations is known to be complex \cite{goyal2022falte, akoury2020storium}. 
Our study aligns with these findings, although focuses on long-form SPTG outputs.

\begin{table}
\caption{\revision{Average Krippendorff's alpha agreement scores between the automatic evaluation metrics and the decisive human annotations, where the annotators explicitly selected $T_+$ or $T_-$.}}\label{tab:krip_alpha}
\centering
\small
\begin{tabular}{lc}
\toprule
 Metric & $\alpha$\\
\midrule
\texttt{BLEU} & 0.0452 $\pm$ 0.0341 \\
\texttt{ROUGE-1} & 0.0459 $\pm$ 0.0281 \\
\texttt{ROUGE-2} & 0.0783 $\pm$ 0.0379 \\
\texttt{ROUGE-L} & \textbf{0.1028} $\pm$ 0.0374 \\
\texttt{METEOR} & 0.0127 $\pm$ 0.0150 \\
\texttt{Wegmann} & 0.0047 $\pm$ 0.0097 \\
\texttt{StyleDistance} & 0.0532 $\pm$ 0.0445 \\
\texttt{Ministral-3B} & 0.0056 $\pm$ 0.0024 \\
\texttt{Llama-3.1-8B} & 0.0152 $\pm$ 0.0303 \\
\texttt{Mistral-24B} & -0.0029 $\pm$ 0.0059 \\
\texttt{Qwen3-32B} & 0.0280 $\pm$ 0.0206 \\
\texttt{DeepSeek-V3} & 0.0472 $\pm$ 0.0166 \\
\texttt{o4-mini} & 0.0618 $\pm$ 0.0425 \\
\texttt{gpt-4.1} & 0.0777 $\pm$ 0.0092 \\
\texttt{$\rho_{all}^{PWV}$} & \underline{0.0842} $\pm$ 0.0389 \\
\bottomrule
\end{tabular}
\end{table}

\revision{
While the collected annotations were highly skewed toward indecisive labels, we still compared human annotations with automatic evaluation metrics by computing Krippendorff’s $\alpha$ between metric predictions and human judgments. To ensure meaningful comparisons, we restricted this analysis to decisive instances, where annotators explicitly selected one of the two candidate outputs.
}
\revision{
Due to the small number of such instances\footnote{The three annotators selected decisive responses for 91, 35, and 43 instances respectively out of the 396 total instances from the first annotation study.}, average agreement between automatic metrics and human judgments was generally low. Nevertheless, the relative agreement trends largely mirror the findings from the main results in Tables~\ref{tab:base_results} and~\ref{tab:ensemble_results_main}, with the exception of \texttt{ROUGE-L}, which exhibits higher agreement than the ensemble $\rho_{all}^{PWV}$ (see Table~\ref{tab:krip_alpha}).
}

\begin{figure}[t]
    \centering
    \begin{subfigure}{\linewidth}
        \centering
        \includegraphics[width=\linewidth]{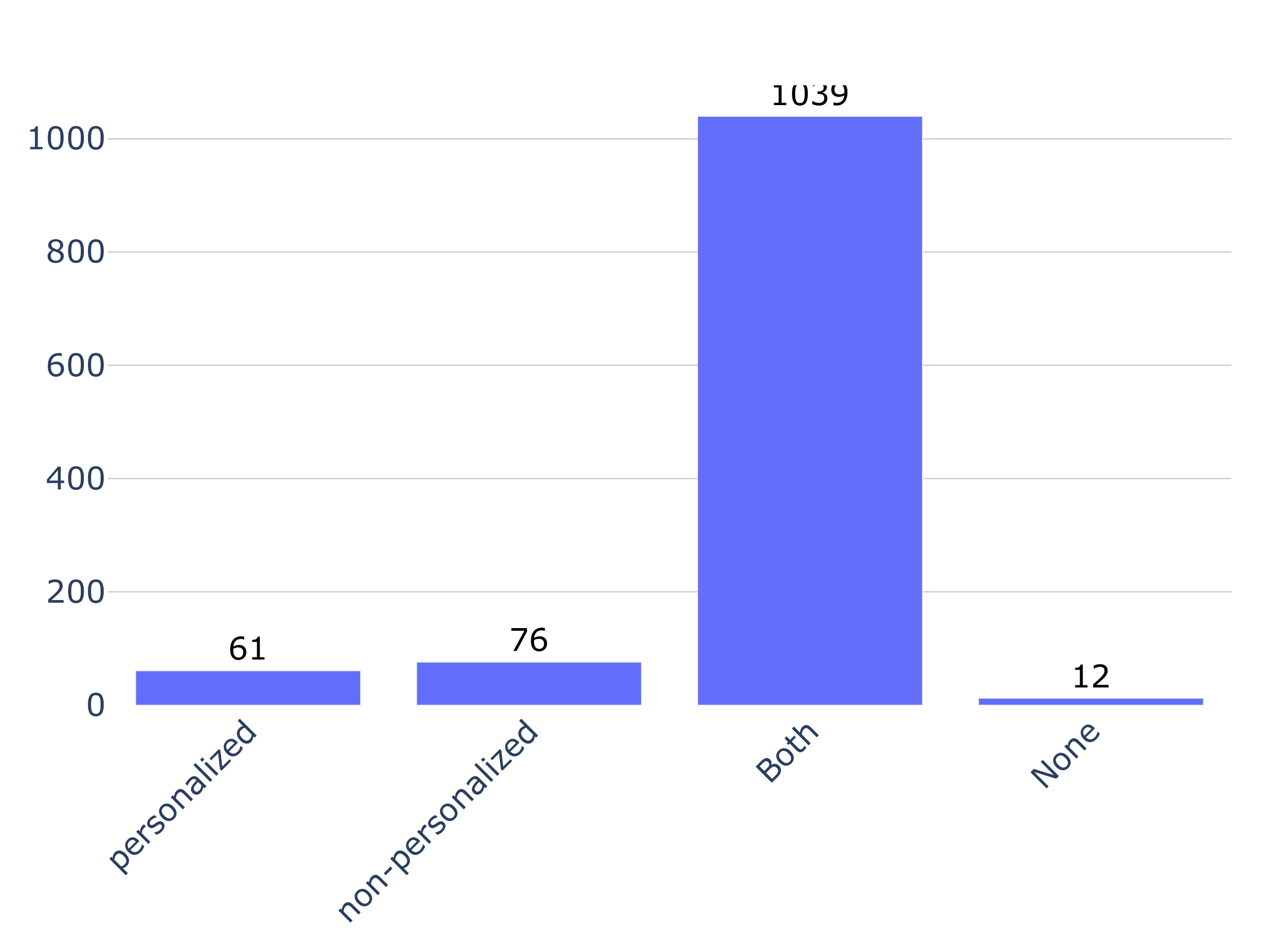}
        \caption{Response distribution for the content preference question ``\textit{Which generated output is better able to cover the content being asked by the |User Input|?}''}
    \end{subfigure} \\
    \begin{subfigure}{\linewidth}
        \centering
        \includegraphics[width=\linewidth]{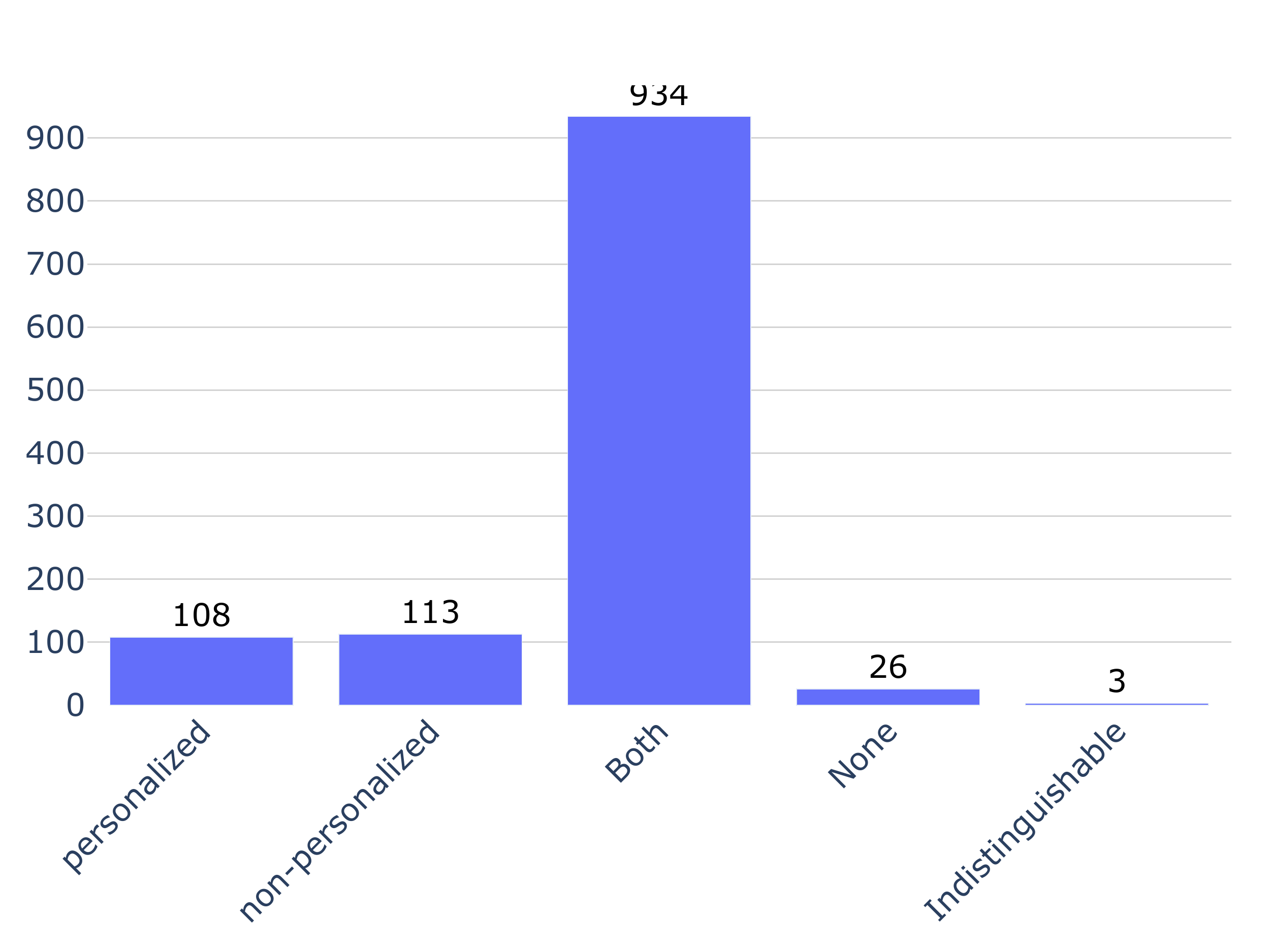}
        \caption{Response distribution for the style preference question ``\textit{Which generated output is better able to reflect the writing style of |Grounding Text|?}''}
    \end{subfigure} \\
    \begin{subfigure}{\linewidth}
        \centering
        \includegraphics[width=\linewidth]{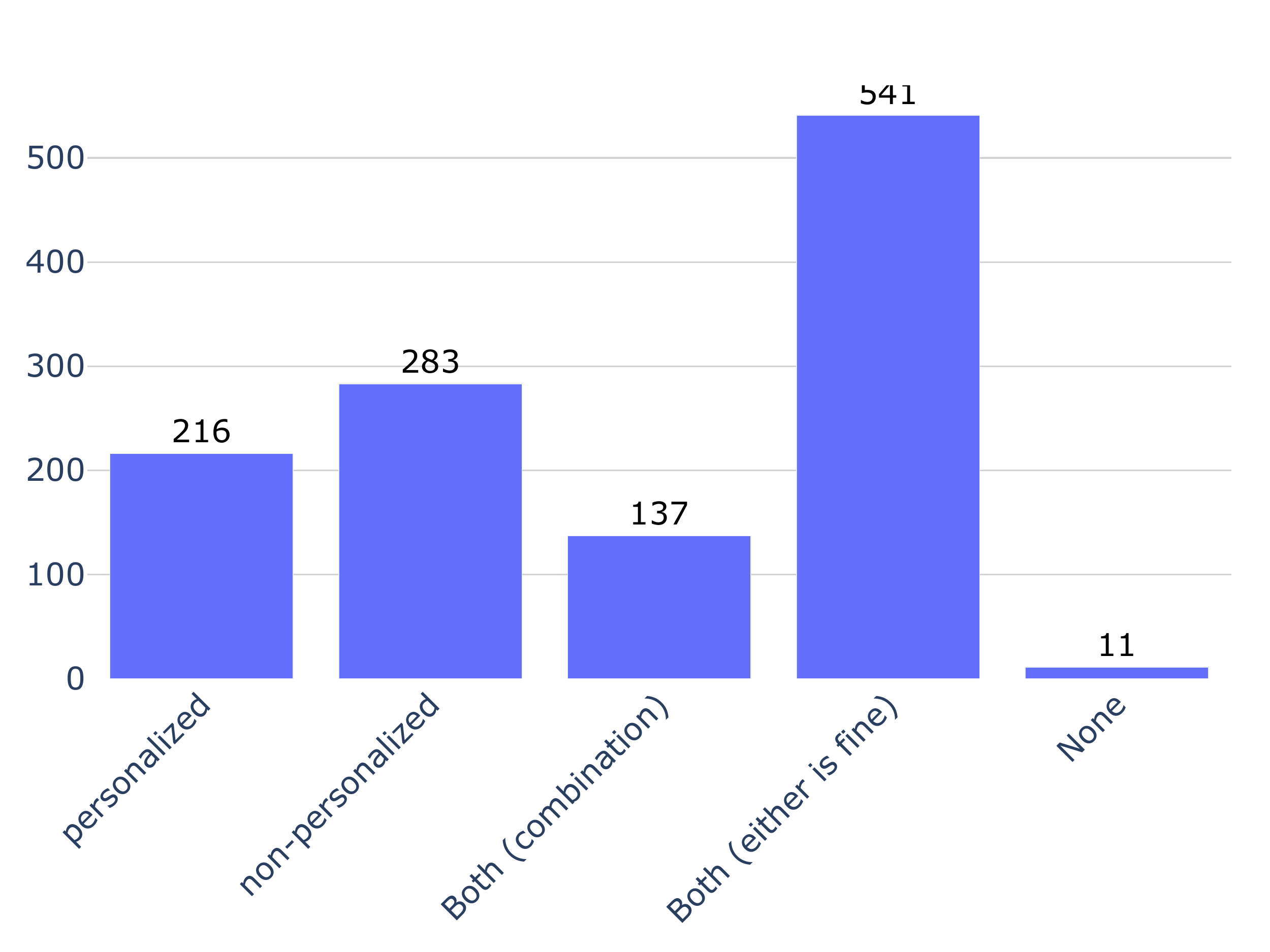}
        \caption{Response distribution for the copy-editing preference question ``\textit{If you were asked to write a response towards |User Input|, which generated output(s) would you use to develop an answer?}''}
    \end{subfigure}
    
    \caption{Distribution of annotation responses for content, style, and copy-editing preference questions.}
    \label{fig:human_eval}
\end{figure}

\begin{figure}[h]
	\centering
	\includegraphics[width=\linewidth]{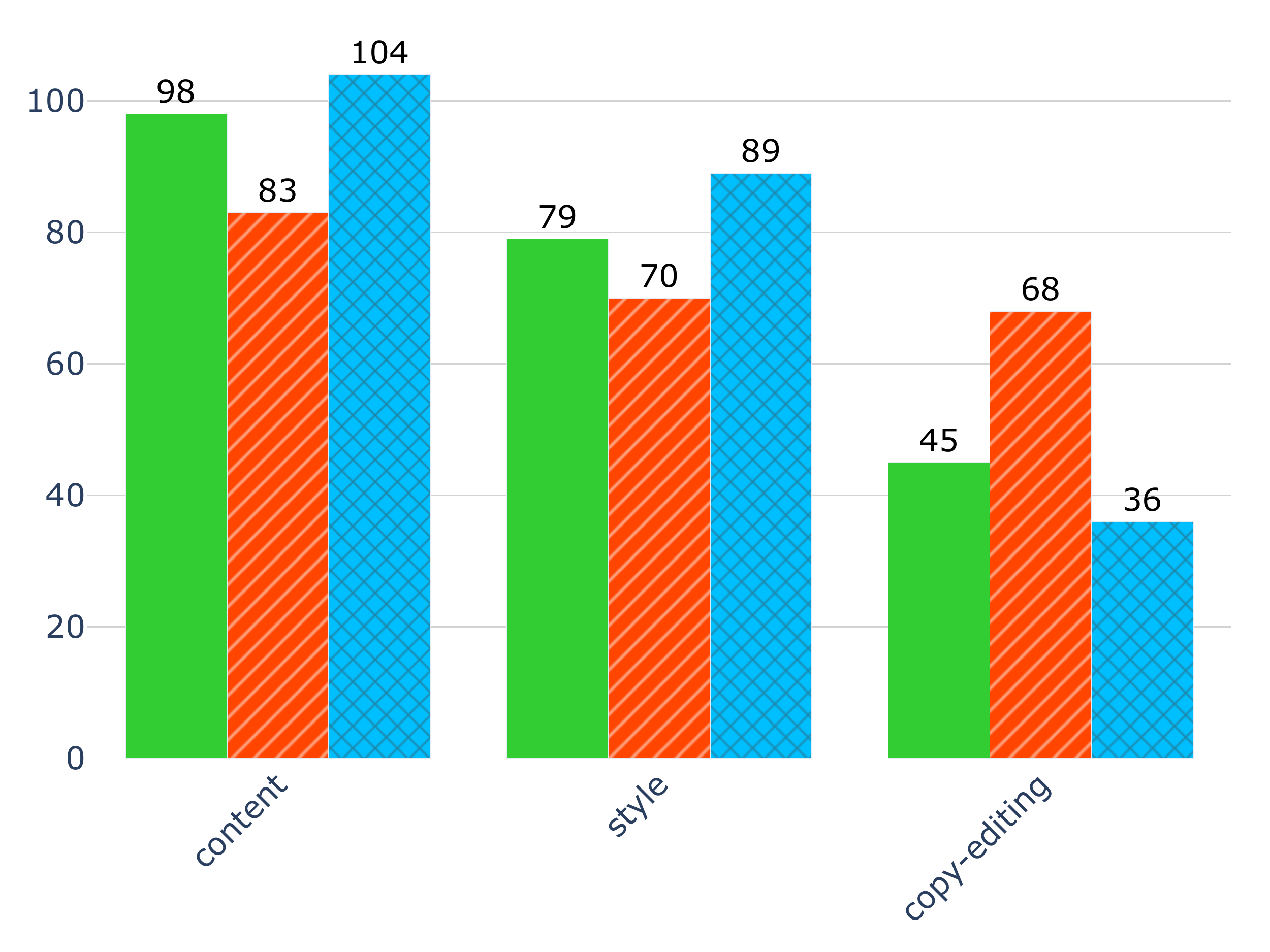}
	\caption{Agreement statistics for three annotators for the style, content and copy-editing preferences for the reproducibility study (out of 129 instances).}
	\label{fig:annotation_reproducibility}
\end{figure}

\paragraph{Second Part: Reproducibility Study}
Given the agreement scores and the takeaways on annotation difficulty, we carried out a second annotation round in order to evaluate the reproducibility of our work. 
For this, we conducted another study with the same annotators one month after the first. 
We used the same annotators under the rationale that, although none of the annotators had authored the texts, they would be most familiar with the existing corpus. 
This study was conducted over a smaller subset of the original corpus, comprising 129 instances. 
We provide the agreement statistics in Figure \ref{fig:annotation_reproducibility}. 
Same as the first study, we found that the annotators were most confident in their content preferences (0.736 IAA) than on the style preferences (r. 0.615). 
Likewise, the copy-editing preferences had the lowest agreement, with about 0.385 agreement across the two annotation studies. 

\end{document}